%% file: main.tex
\documentclass[10pt,twocolumn,letterpaper]{article}

\usepackage[pagenumbers]{cvpr} 

\input{preamble}
\usepackage{bbm}
\definecolor{cvprblue}{rgb}{0.21,0.49,0.74}
\usepackage[pagebackref,breaklinks,colorlinks,allcolors=cvprblue]{hyperref}

\def\confName{CVPR}
\def\confYear{2026}

\title{Ar2Can: An Architect and an Artist Leveraging \\ a Canvas for Multi-Human Generation}

\author{Shubhankar Borse$^\S$ \quad Phuc Pham \quad Farzad Farhadzadeh \quad Seokeon Choi$^\S$ \\
Phong Nguyen \quad Anh Tran \quad Sungrack Yun \quad Munawar Hayat$^\S$ \quad Fatih Porikli\\
Qualcomm AI Research\thanks{Qualcomm AI Research, an initiative of Qualcomm Technologies, Inc.} \\
$^\S${\tt\small \{sborse, seokchoi, mhayat\}@qti.qualcomm.com}
}

\begin{document}
\maketitle

\input{sec/0_abstract}

\input{sec/1_intro}

\input{sec/2_related}

\input{sec/3_method}
\input{sec/4_results}


{
    \small
    \bibliographystyle{ieeenat_fullname}
    \bibliography{main}
}

\clearpage

\input{sec/X_suppl}

\end{document}

%% file: preamble.tex
\usepackage{algorithm}
\usepackage{algpseudocode}
\usepackage{makecell}
\usepackage{multirow}
\usepackage{appendix}
\usepackage{array,booktabs,multirow,colortbl,xcolor}
\definecolor{lightyellow}{RGB}{255,250,205}
\definecolor{lightpurple}{RGB}{220,220,255}
\definecolor{lightblue}{RGB}{230,240,255}
\definecolor{lightred}{RGB}{255,230,230}
\definecolor{lightgreen}{RGB}{230,255,230}
\definecolor{darkyellow}{RGB}{153,102,0}
\usepackage{etoc}
\usepackage[accsupp]{axessibility}

\definecolor{Gray}{gray}{0.92}

\usepackage{minted}



%% file: sec/0_abstract.tex

\begin{abstract}

Despite recent advances in personalized image generation, existing models consistently fail to produce reliable multi-human scenes, often merging or losing facial identity. We present Ar2Can, a novel two-stage framework that disentangles spatial planning from identity rendering for multi-human generation. The Architect predicts structured layouts, specifying where each person should appear. The Artist then synthesizes photorealistic images, guided by a spatially-grounded face matching reward that combines Hungarian spatial alignment with identity similarity. This approach ensures faces are rendered at correct locations and faithfully preserve reference identities. We develop two Architect variants, seamlessly integrated with our diffusion-based Artist model. This is optimized via Group Relative Policy Optimization (GRPO) using compositional rewards for count accuracy, image quality, and identity matching. Evaluated on the MultiHuman-Testbench, Ar2Can achieves substantial improvements in both count accuracy and identity preservation, while maintaining high perceptual quality. Notably, our method achieves these results using primarily synthetic data, without requiring real multi-human images. Project page: \url{https://qualcomm-ai-research.github.io/ar2can/}.
\vspace{-0.6cm}
\end{abstract}

%% file: sec/1_intro.tex

\begin{figure}[t]
  \centering
  \includegraphics[width=\linewidth]{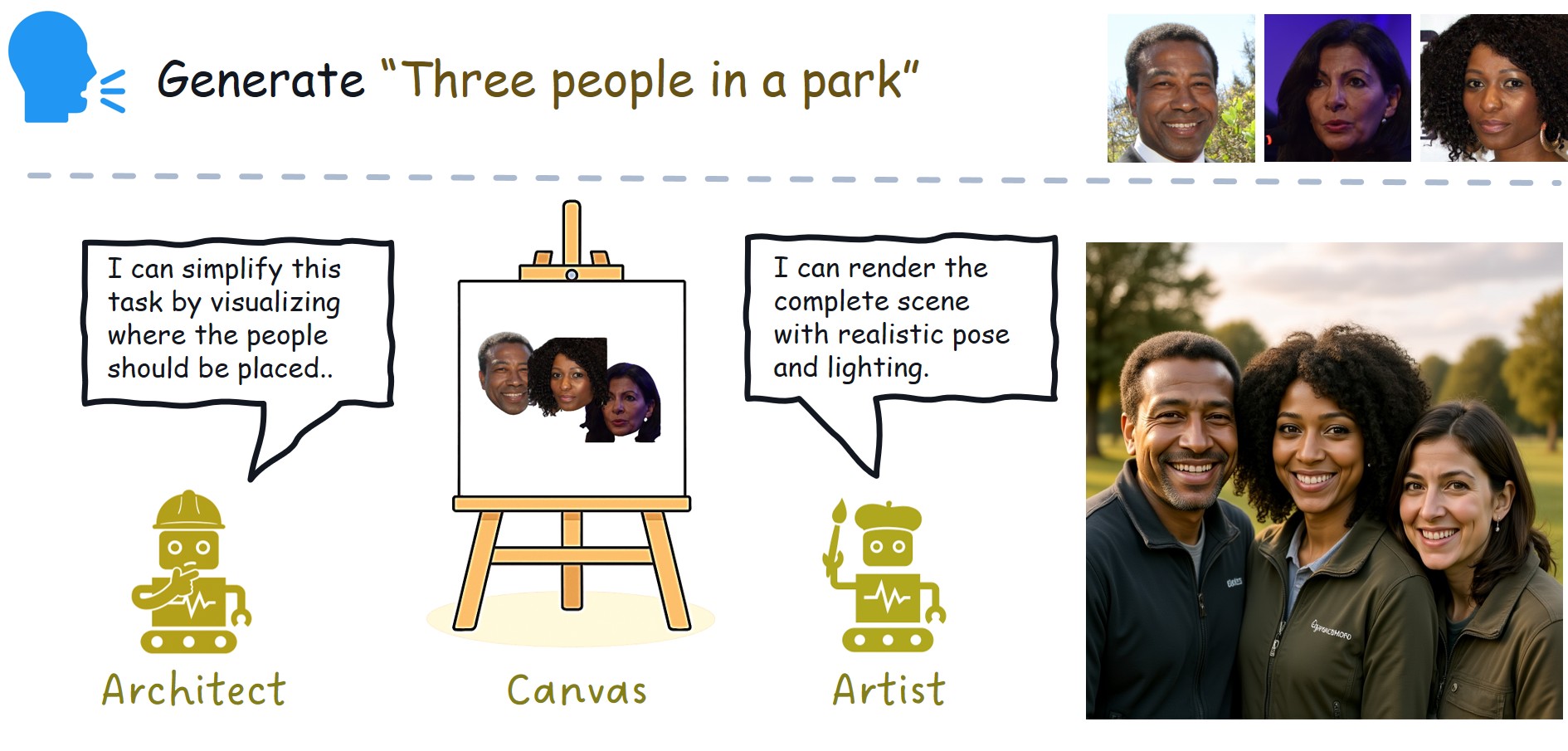}
  \caption{\textbf{Ar2Can Framework Overview.} Our two-stage approach decomposes multi-human generation into spatial planning (Architect) and identity-preserving rendering (Artist).}
  \vspace{-1 em}
  \label{fig:teaser}
\end{figure}

\section{Introduction}
\label{sec:intro}
Text-to-image diffusion models~\cite{rombach2022high, podell2023sdxl, flux2024} have achieved photorealistic synthesis quality. However, they systematically fail on a fundamental task: generating scenes with multiple distinct humans~\cite{borse2025multihuman}. Given a prompt such as ``five friends having coffee'' and reference identity images, current approaches produce images with duplicated faces, merged identities, or incorrect person counts (Fig.~\ref{fig:results_comparison}). This failure persists across different paradigms: regional conditioning methods~\cite{li2023gligen, xu2025withanyone, zhang2025id} require explicit spatial annotations at inference time, limiting usability; identity-preserving approaches~\cite{guo2024pulid, labs2025flux1kontext} excel at single-subject personalization but struggle with multi-identity composition; and even methods explicitly designed for multi-ID generation~\cite{wu2025omnigen2, xiao2025omnigen, mou2025dreamo, chen2025xverse, cheng2025umo} fail on recently introduced benchmarks~\cite{borse2025multihuman} (Section~\ref{sec:experiments}). This failure is a fundamental limitation in the way models represent and compose multiple identities within the same scene.

\begin{figure*}[t]
\centering
\includegraphics[width=0.85\textwidth]{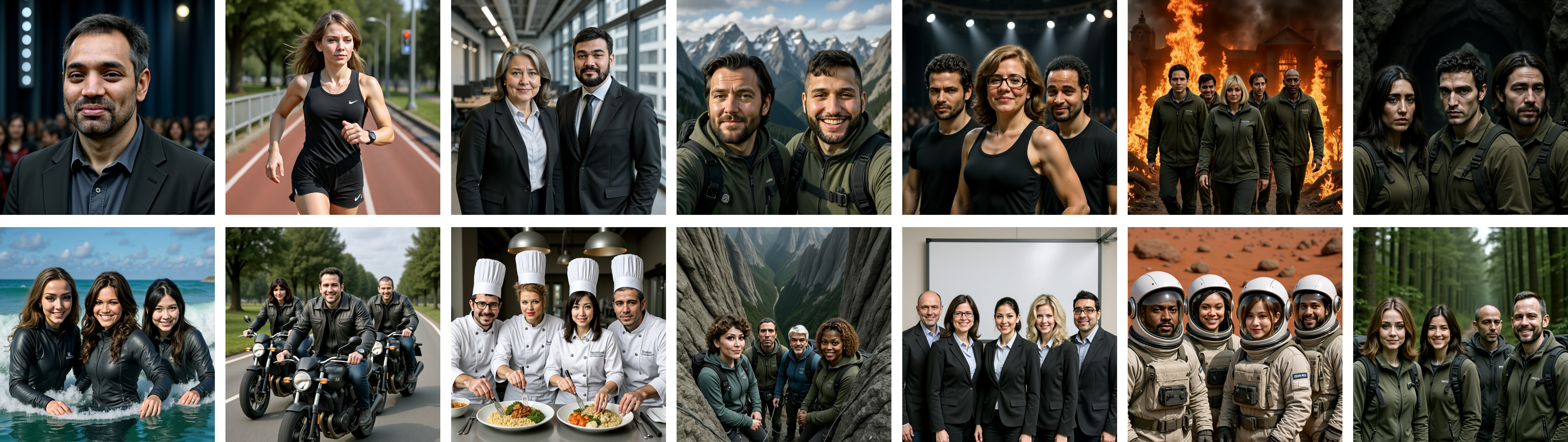}
\caption{Ar2Can generates highly photorealistic multi-human scenes with 1-5 people while preserving the individual identities. Our two-stage architecture produces natural poses, realistic lighting, and proper spatial arrangements without identity merging or blending artifacts.}
\label{fig:results_gallery}
\vspace{-0.7 em}
\end{figure*}

The key insight motivating our work is that current methods \textit{merge spatial layout and identity rendering in a single-stage generation process}. When a diffusion model simultaneously reasons about ``where people should be'' and ``what they should look like,'' it conflates spatial structure with appearance, leading to systematic failures. Consider generating three people: without explicit spatial constraint, the model may place multiple faces in overlapping regions causing identity merging, or generate the same face in different locations causing duplication.

\noindent \textbf{Our Approach.}
We propose \textbf{Ar2Can} (\textit{Architect, Artist and Canvas}), a two-stage framework that disentangles spatial planning from identity rendering. As illustrated in \cref{fig:teaser}, the \textbf{Architect} first generates a structured spatial plan: bounding boxes (and optionally pose) specifying where (and how) each person should appear. We design two Architect variants with complementary trade-offs: \textbf{Architect-A} uses Qwen~\cite{qwen2024} with supervised fine-tuning on bounding box prediction (L1 + GIoU losses), providing structured spatial reasoning; \textbf{Architect-B} uses Flux-Schnell~\cite{flux2024} finetuned with reinforcement learning, enabling adaptive planning with human pose. 

The \textbf{Artist} then renders the photorealistic image conditioned on this spatial plan. We train a single Artist model, \textbf{Flux-Kontext}, via Group Relative Policy Optimization (GRPO)~\cite{shao2024deepseekmath} that works with either Architect variant. Critically, we introduce a \textit{spatially-grounded face matching reward} that operates in two stages: (1) Hungarian algorithm matching between expected centroids (from the Architect) and detected face locations, and (2) ArcFace~\cite{deng2019arcface} identity similarity between reference faces and their Hungarian-matched generated counterparts. Combined with count accuracy, pose correction and HPSv3 quality rewards~\cite{ma2025hpsv3}, this compositional reward structure ensures faces appear at correct locations \textit{and} match the right identities. 

\noindent \textbf{Contributions.}
\begin{enumerate}[leftmargin=*, nosep]
  \item We propose Ar2Can, a two-stage framework which disentangles spatial planning (Architect) from identity rendering (Artist) for personalized multi-human generation.
  
  \item We propose two Architect variants (supervised Qwen and. RL-finetuned Schnell) with different speed, accuracy and alignment trade-offs.
  
  \item We propose an RL-based Artist training method using compositional rewards: hungarian based spatially-grounded face matching, prompt and pose alignment, and perceptual quality. We combined it with curriculum learning to jointly optimize spatial accuracy, identity preservation, and photorealism while preventing copy-paste artifacts.

  \item We introduce a token sharing and dropping strategy that reduces inference time by 2× while empowering the model to implicitly handle occlusion cases through shared positional encodings.

  \item Evaluated on MultiHuman-Testbench~\cite{borse2025multihuman}, Ar2Can demonstrates substantial improvements in count accuracy and identity preservation over state-of-the-art methods while maintaining competitive perceptual quality.
\end{enumerate}

%% file: sec/2_related.tex
\section{Related Work}
\label{sec:related}
\vspace{-0.1em}
\noindent \textbf{Text-to-Image Diffusion Models.}
Recent diffusion models~\cite{rombach2022high, saharia2022photorealistic, ramesh2022hierarchical, podell2023sdxl, esser2024scaling, flux2024} achieve remarkable realism but struggle with compositional tasks~\cite{chefer2023attend, liu2022compositional}, particularly multi-human generation~\cite{borse2025multihuman, disco2025}. To address spatial composition, methods such as GLIGEN~\cite{li2023gligen}, ReCo~\cite{yang2023reco}, LayoutDiffusion~\cite{zheng2023layoutdiffusion}, and BoxDiff~\cite{xie2023boxdiff} introduce \textbf{regional control} via bounding boxes, but require explicit user annotations at inference time. LayoutGPT~\cite{feng2023layoutgpt} and RPG~\cite{yang2024mastering} automate layout generation with off-the-shelf LLMs, but prove suboptimal for multi-human scenes (Appendix~\ref{app:offtheshelf_arch}). Unlike these, we fine-tune specialized Architects that generate accurate, prompt-aware layouts without user intervention.

\noindent \textbf{Identity-Preserving and Multi-Human Generation.}
IP-Adapter~\cite{ye2023ip}, InstantID~\cite{wang2024instantid}, PuLID~\cite{guo2024pulid} and others~\cite{qian2025omni, zhou2024storymaker, jiang2025infiniteyou, he2025uniportrait} enable high-fidelity identity injection but are limited to single-subject personalization. Multi-human approaches spanning identity-preserving diffusion~\cite{labs2025flux1kontext}, regional conditioning~\cite{xu2025withanyone, zhang2025id, huang2025resolving}, and unified architectures~\cite{wu2025omnigen2, xiao2025omnigen, mou2025dreamo, chen2025xverse, cheng2025umo, qian2025composeme} all exhibit systematic failures on MultiHuman-Testbench~\cite{borse2025multihuman}. Concurrent canvas-based methods~\cite{qian2025layercomposer, dalva2025canvas} share our spatial intuition but rely on large proprietary multi-human datasets, whereas Ar2Can achieves superior performance using only synthetic multi-human data and sophisticated RL finetuning.

\noindent \textbf{Reinforcement Learning for Diffusion.}
DDPO~\cite{black2023training} introduced policy-gradient RL for diffusion models, with learned reward functions~\cite{xu2023imagereward, ma2025hpsv3} and extensions to diverse objectives~\cite{fan2023dpok, prabhudesai2023aligning} following. DisCo~\cite{disco2025} applied Flow-GRPO~\cite{liu2025flow} to multi-human generation with diversity rewards. Ar2Can extends this paradigm with novel spatially-grounded rewards that explicitly couple face locations with planned layouts, directly resolving identity merging and swapping failures absent in prior RL-based approaches.

%% file: sec/3_method.tex

\section{Method}
\label{sec:method}
\subsection{Overview}
\label{sec:overview}
\vspace{-0.5em}
Ar2Can decomposes multi-human image generation into two stages (\cref{fig:teaser}). Given a text prompt $p$ (e.g., ``three people playing basketball'') and $N$ reference identity images $\{I_{\text{ref},1}, \ldots, I_{\text{ref},N}\}$, the \textbf{Architect} module $\psi$ predicts a spatial layout $\mathcal{L} = \{b_1, \ldots, b_N\}$, where each bounding box $b_i = (x_i, y_i, w_i, h_i)$ defines a region with centroid $c_i = (x_i + w_i/2, y_i + h_i/2)$. Conditioned on the prompt, layout, and reference images, the \textbf{Artist} module $\pi_\theta$ synthesizes the final image via $x \sim \pi_\theta(x \mid p, \mathcal{L}, \{I_{\text{ref}}\})$. This decomposition offers two key advantages: (1) \textit{Explicit spatial grounding:} The face locations are determined prior to rendering, reducing spatial conflicts that can lead to identity merging. (2) \textit{Modular design:} the Architect module can be replaced based on deployment constraints (e.g., speed vs. accuracy) without requiring retraining of the Artist. This enables us to benchmark methods such as WithAnyone~\cite{xu2025withanyone} and ID-Patch~\cite{zhang2025id} using our proposed architects. Theoretically, this decomposition reduces error propagation: since our Architects achieve high standalone layout accuracy (Appendix~D), the Artist can focus solely on identity-preserving rendering rather than compensating for spatial errors, as confirmed in Section~\ref{sec:quantitative}.

\subsection{Training Data Curation}
\label{sec:data_curation}



A central challenge in multi-human image generation is the limited availability of supervised training data. Ideally, one would fine-tune models on millions of images featuring multiple individuals within the same scene, each accompanied by distinct reference images representing their identities. Such large-scale datasets are typically proprietary, confined to organizations with access to vast user-generated content(such as Snap or Meta). Public datasets for Multi-Human generation(such as LAION-Face, Multi-ID-2M~\cite{xu2025withanyone}) contain lots of images sampled from videos, which aren't high quality. There are very few images of ``Portrait-style, High-quality" open source data; hence methods such as~\cite{zhang2025id} curate datasets by purchasing data. However, rather than relying on this scarce real multi-person data, we leverage DisCo~\cite{disco2025} to generate synthetic multi-person scenes, then construct hybrid training samples by pairing these scenes with real reference faces. Instead of supervising our \textbf{Artist} with unavailable data, we \textbf{reinforce} its predictions using this curated data. This helps to ground the model in realistic facial appearances and also capitalizes on DisCo's compositional capabilities.

\noindent \textbf{Data Sources.}
We collect training samples from three sources, which offered varying levels of identity supervision.

\begin{enumerate}[leftmargin=*, nosep]
  \item \textbf{Multi-view references} ($\sim$100K identities): Public datasets consisting of individuals with an additional view from a different angle~\cite{guo2016ms, xu2025withanyone}, enabling multi-view identity learning.
  
  \item \textbf{Single-view references} ($\sim$500K images): Public datasets with single high-quality images lacking a second view~\cite{yang2024consistentid}. For a percentage of these, we synthetically generate secondary views using PuLID~\cite{guo2024pulid} to introduce diversity in pose and expressions. For the rest, we apply random augmentations including rotations, horizontal flipping, and brightness adjustment.
   
  \item \textbf{Synthetic faces}: Identity-varied faces already generated during scene synthesis by DisCo. These enhance robustness through synthetic diversity.
\end{enumerate}

\noindent \textbf{Canvas Construction Pipeline.}
We construct each training sample via a four-stage pipeline, which produces hybrid samples combining synthetic multi-person scenes with real reference faces. A detailed Algorithm and visuals are present in Appendix C. The process begins by generating a DisCo scene with $N$ people and detecting face locations. Then, we initialize a blank canvas and replace each synthetic face with a reference face from our curated sources. Each training sample comprises three components: the constructed canvas, the original reference face images, and the DisCo-generated target image. Our ``pose-controlled'' Artist(discussed in Appendix C) also pastes the pose for each human on the canvas. Our dataset statistics and hyperparameters are discussed in Appendix B. 

\subsection{Architect: Spatial Layout Generation}
\label{sec:architect}
In this section, we explore Architect designs aimed at generating facial bounding boxes and/or pose from textual descriptions, with a focus on achieving accurate instance counts and spatially plausible placements.

\subsubsection{Architect-A: Bounding-Box Co-ord. Regression}
\label{sec:architect_a}



Directly regressing bounding-box coordinates from text prompts is suboptimal, as it faces two challenges: (\textit{i}) the number of boxes varies dynamically with the prompt, and (\textit{ii}) extracting counts requires deep linguistic reasoning (e.g., ``Two girls next to a man and behind a lady''). We address this issue by building Architect-A as an autoregressive model using Qwen-2.5 (0.5B)~\cite{qwen2024}, which provides both dynamic output length and inherent language understanding. 

\begin{figure}
    \centering
    \includegraphics[width=1\linewidth]{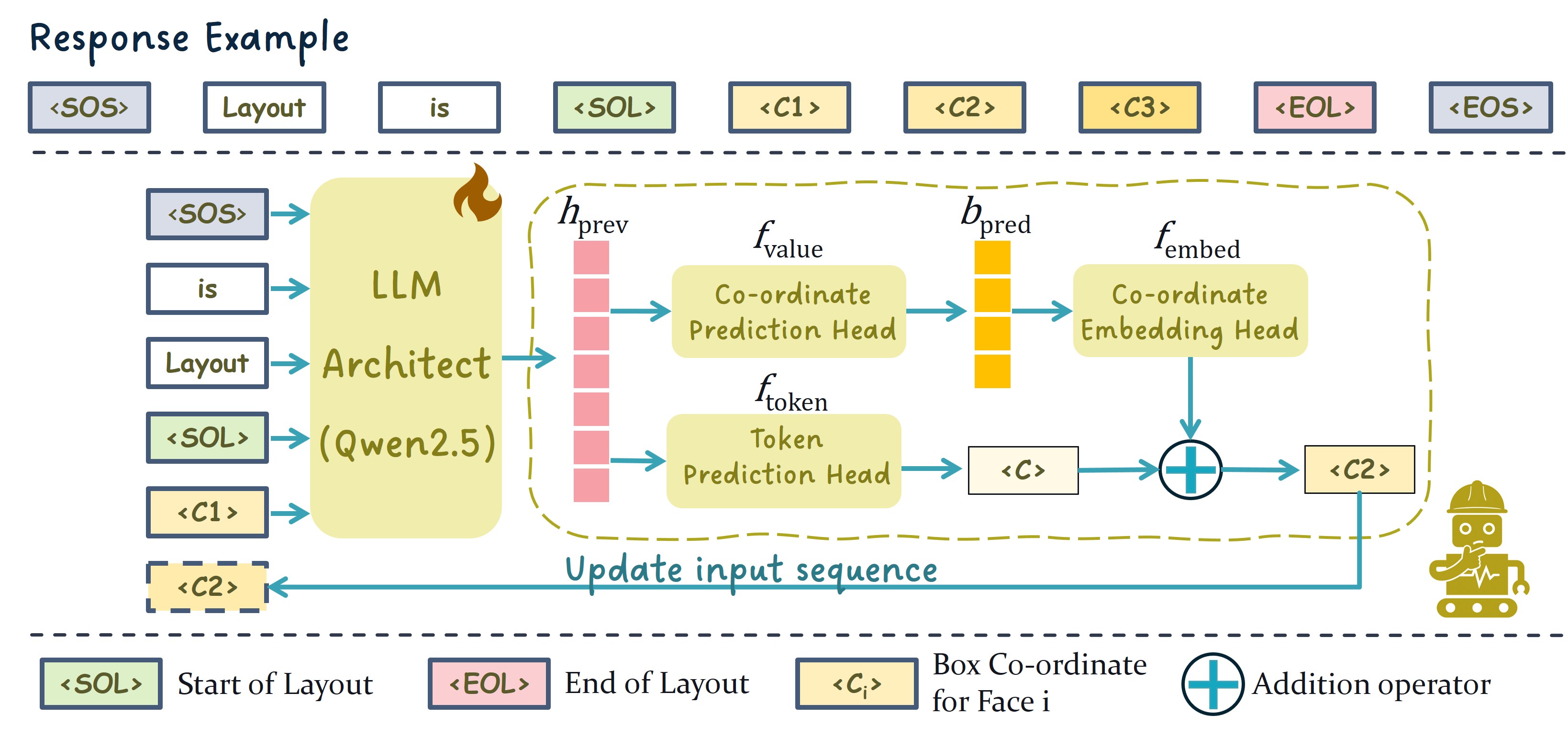}
    \caption{Architecture of the LLM-based Architect-A for layout. \textbf{Top}: response example for spatial layout generation. \textbf{Bottom}: our lightweight LLM extended with special tokens.} 
    \label{fig:qwen-arch}
\end{figure}

Figure~\ref{fig:qwen-arch} illustrates our Architect-A training pipeline. As observed, we extend the tokenizer with layout structure tokens: \texttt{<SoL>} (start of layout), \texttt{<EoL>} (end of layout) and \texttt{<C>} (coordinate placeholder). Since coordinates require continuous values that standard LLMs cannot directly output, we augment the model with additional prediction heads. Following prior work~\cite{lisa, chatpose, aipparel, chatgarment}, we attach two heads to the base LLM: a coordinate regression head $f_{\text{value}}$ and a coordinate embedding head $f_{\text{embed}}$, alongside the original token head $f_{\text{token}}$. 

When $f_{\text{token}}$ predicts a \texttt{<C>} token, $f_{\text{value}}$ is triggered to output its corresponding continuous coordinate values. Since the model must output multiple bounding boxes (one per person), multiple \texttt{<C>} tokens appear in the sequence. We re-embed each predicted coordinate value with $f_{\text{embed}}$ to form instance-specific tokens $\texttt{<C}_\texttt{i}\texttt{>}$, where $i$ is the person index. This  differentiates coordinate tokens for different individuals, improving spatial consistency and reducing ambiguity during the autoregressive generation process.

\noindent \textbf{Training Dynamics}
We optimize Architect-A with two objectives: cross-entropy loss $\mathcal{L}_{\text{CE}}$ for token prediction and coordinate loss $\mathcal{L}_{\text{coord}}$ for box regression:
\begin{align}
\mathcal{L}_{\text{Arch-A}} &= \mathcal{L}_{\text{CE}} + \lambda_{\text{coord}} \mathcal{L}_{\text{coord}} \nonumber \\
&= \text{CE}(f_{\text{token}}(h_{\text{prev}}), y_{\text{next}}) \nonumber \\ & + \lambda_{\text{coord}} \mathbb{E}_{b_{\text{gt}}} 
\left[
\mathcal{L}_{\text{gIoU}}(b_{\text{pred}}, b_{\text{gt}})
+ \|b_{\text{pred}} - b_{\text{gt}}\|_1
\right] 
\end{align}

where $b_{\text{pred}} = f_{\text{value}}(h_{\text{prev}})$ are predicted coordinates from LLM hidden states $h_{\text{prev}}$, $y_{\text{next}}$ denotes the next token to be predicted, and $b_{\text{gt}}$ are ground-truth boxes. We use gIoU~\cite{rezatofighi2019generalized} to handle non-overlapping boxes and L1 loss for stable gradients with normalized coordinates. We also observed that MLP layers are highly sensitive to permutation of bounding box coordinate embeddings, negatively affecting performance and convergence. Hence, we enforce consistent coordinate ordering via \textbf{data sorting}. This simple strategy improves training stability and enables more reliable coordinate regression. Additional implementation details and explanations are presented in the Appendix.



\subsubsection{Architect-B: T2I-based Layout Generation}
\label{sec:architect_b}

Direct bounding-box regression offers simplicity. However, in some cases, it might not capture complex spatial relationships. Hence, an alternative to this method is to leverage text-to-image (T2I) generation, which operates in 2D and thus provides richer spatial priors. In practice, a fast and high-quality T2I model can synthesize a draft image, from which a plausible layout can be extracted. A key challenge is ensuring accurate person counts, as standard T2I models frequently produce incorrect counts in multi-human scenes. This limitation can be addressed by fine-tuning the base T2I model via reinforcement learning, aligning the generated content with the desired count.


We adopt Flux-Schnell~\cite{flux2024} as the base T2I model for layout sketching. It is a widely used flow-matching model that achieves high speed by requiring only 4 denoising steps. However, it struggles with accurate person count generation and action alignment in multi-human scenes (see Appendix D). To address these limitations, we fine-tune Flux-Schnell using Group Relative Policy Optimization (GRPO), using a setup similar to DisCo~\cite{disco2025}. It recently leveraged Flow-GRPO~\cite{liu2025flow} to demonstrate an effective RL Framework for improving count accuracy (and diversity) in multi-human generation.

\noindent \textbf{Training Dynamics:}
GRPO computes group-normalized advantages over $M$ samples per prompt: $A_i = (r(x_i, p) - \mu_G) / (\sigma_G + \epsilon)$ where $\mu_G, \sigma_G$ are the mean and standard deviation of rewards within group $G$, and $\epsilon$ prevents division by zero. The policy $\pi_\theta$ is optimized via:
\begin{equation}
\mathcal{L}_{\text{GRPO}} = \mathbb{E}_{p, G} \left[ \sum_{i=1}^M A_i \log \frac{\pi_\theta(x_i|p)}{\pi_{\text{ref}}(x_i|p)} - \beta_{\text{KL}} \text{KL}(\pi_\theta || \pi_{\text{ref}}) \right]
\end{equation}
where $\pi_{\text{ref}}$ is the frozen reference policy (base Flux-Schnell) and $\beta_{\text{KL}}$ controls policy divergence. We guide the fine-tuning with a compositional reward combining count accuracy and prompt alignment:
\begin{equation}
r_{\text{Arch-B}}(x, p) = \alpha \cdot r_{\text{count}}(x) + \beta \cdot r_{\text{hps}}(x, p)
\end{equation}
where $r_{\text{count}}(x) = \mathbbm{1}[n_{\text{pred}}(x) = n_{\text{target}}]$ rewards exact person count matching (detected via blob analysis), and $r_{\text{hps}}(x, p) = \text{HPSv3}(x, p)$ measures prompt alignment using Human Preference Score v3~\cite{ma2025hpsv3}. It prevents spatial reward hacking in Architect-B and improves the spatial arrangement of faces relative to the prompt (Appendix~D). The fine-tuned Schnell model generates both face bounding boxes and human pose coordinates for downstream use by the Artist. Training hyperparameters are detailed in Appendix B. 

Each Architect design has its own advantages: Architect-A has strong language understanding for accurate counts, while Architect-B produces plausible spatial locations and pose more effortlessly thanks to its 2D nature. We selected both these variants primarily for their \textbf{efficiency tradeoff}. To validate this choice, we additionally explored \textbf{Architect-C}, a unified multimodal architecture (BAGEL~\cite{deng2025emerging}) finetuned using the Arch-B RL procedure. It achieves comparable Artist performance but at significantly higher training and inference cost (Appendix~D). We compare all designs and discuss trade-offs further in Section~\ref{sec:experiments} and Appendix~D.

\begin{figure*}[t]
\centering
\includegraphics[width=0.9\textwidth]{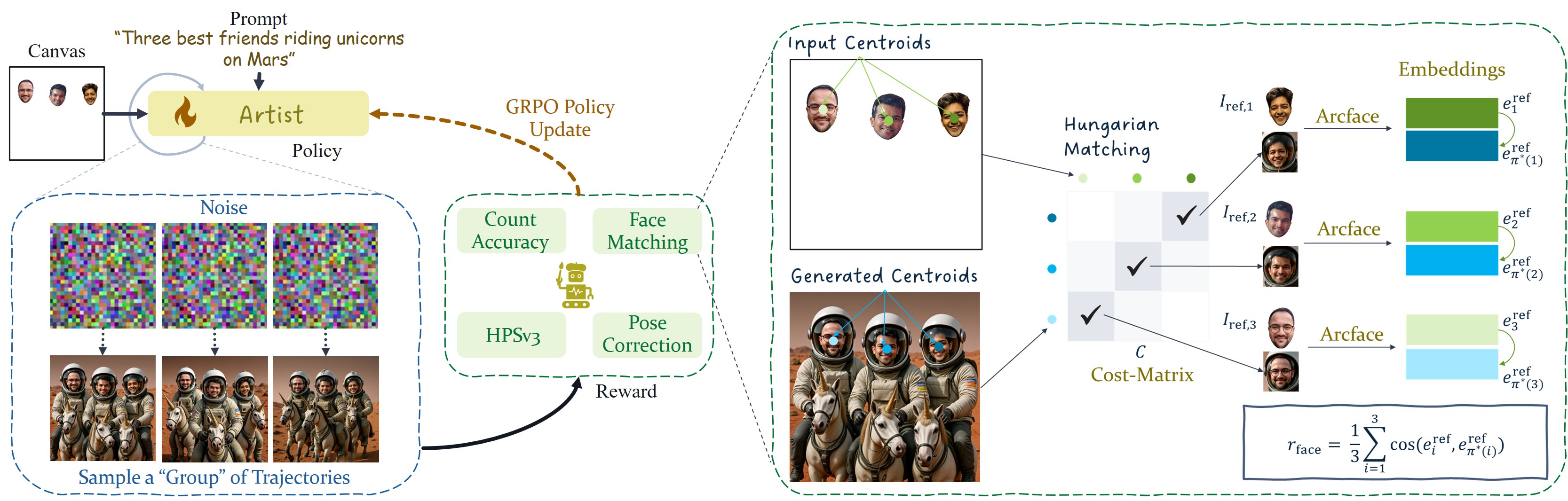}
\caption{\textbf{Artist training pipeline with GRPO.} Given the input canvas and text prompt, sample a group of images and optimize over compositional rewards: count accuracy, prompt alignment/aesthetic quality (HPSv3), spatially-grounded face matching and pose correction.}
\label{fig:artist_training}
\end{figure*}

\subsection{Artist: Identity-Preserving Rendering}
\label{sec:artist}

Using the Architect's boxes, we construct the canvas layout $\mathcal{L}$ (details in Appendix~\ref{app:inference_canvas}). The Artist is a single model, Flux-Kontext, trained to render photorealistic multi-human images conditioned on $\mathcal{L}$, text prompt $p$, and reference images $\{I_{\text{ref}}\}$. While Kontext excels at single-identity preservation, it largely fails in multi-human scenarios. It merges faces into similar identities and swaps reference identities across spatial locations. In many cases, it is not able to retain more than one input identity, as clear in Table~\ref{tab:multihuman_recent}. These failures reflect an implicit bias from pre-training on predominantly single-subject data. To fix this issue, we employ reinforcement learning rather than supervised fine-tuning as: (1) RL enables direct optimization of non-differentiable objectives (e.g. count, identity matching) via off-the-shelf detectors, and (2) supervised approaches would require large-scale annotations pairing reference identities with correct spatial locations. We can bypass this by using the data we curated, as discussed in Section~\ref{sec:data_curation}. Hence, RL allows us to explicitly \textbf{correct} the model's multi-human behavior without expensive annotations.

\subsubsection{GRPO Training with Compositional Rewards}
\label{sec:artist_training}
We train the Artist via GRPO using four compositional rewards:
\begin{align}
\begin{split}
r_{\text{Artist}}(x, p, \mathcal{L})  & =  \, \alpha \cdot r_{\text{count}}(x) + \beta \cdot r_{\text{hps}}(x, p) \\
& + \zeta \cdot r_{\text{face}}(x, \mathcal{L}) + \eta \cdot r_{\text{pose}}(x, \mathcal{L})
\end{split}
\end{align}
where $r_{\text{count}}$ and $r_{\text{hps}}$ are defined as in \cref{sec:architect_b}, $r_{\text{face}}$ is our spatially-grounded face matching reward, and $r_{\text{pose}}$ measures pose alignment between generated humans and pose specifications. We illustrate this training in Figure~\ref{fig:artist_training}. The HPSv3 reward provides a strong capability to adhere to the input prompt, and also heavily enhances realism. 

\noindent \textbf{Face Matching Reward.}
The face matching reward ensures generated faces (1) appear near Architect-specified locations and (2) match reference identities. A naive location-based approach, i.e. directly extracting face crops at predicted bounding boxes, leads to copy-paste artifacts and reward hacking. The model simply pastes reference faces at exact locations with unnatural face sizes due to the lookup area (see Appendix E for visual examples). 

To address this pressing issue, we propose a two-stage matching strategy. This is illustrated in Figure~\ref{fig:artist_training}. Given the Architect's predicted bounding boxes $\{b_1^{\text{pred}}, \ldots, b_N^{\text{pred}}\}$ with centroids $\{c_i^{\text{pred}}\}$ and RetinaFace~\cite{deng2020retinaface} detected boxes $\{b_1^{\text{det}}, \ldots, b_O^{\text{det}}\}$ with centroids $\{c_j^{\text{det}}\}$, we first establish spatial correspondence via Hungarian matching~\cite{kuhn1955hungarian} on centroid distance cost matrix $C_{ij} = \| c_i^{\text{pred}} - c_j^{\text{det}} \|_2$, finding optimal assignment $\pi^* = \arg\min_\pi \sum_{i=1}^{\min(N,O)} C_{i,\pi(i)}$. This centroid-based matching eases the constraint on exact localization, i.e. the Artist must generate faces \textit{near} the specified locations rather than at precise coordinates, enabling photorealistic rendering with natural variations in pose, scale, and positioning. 

For each reference image $I_{\text{ref},i}$, we then compute identity similarity with its spatially-matched generated face using ArcFace~\cite{deng2019arcface} embeddings $e_i^{\text{ref}} = f_{\text{ArcFace}}(I_{\text{ref},i})$ and $e_{\pi^*(i)}^{\text{gen}} = f_{\text{ArcFace}}(\text{crop}(x, b_{\pi^*(i)}^{\text{det}}))$:
\begin{equation}
s_i = \begin{cases}
\frac{e_i^{\text{ref}} \cdot e_{\pi^*(i)}^{\text{gen}}}{\| e_i^{\text{ref}} \| \| e_{\pi^*(i)}^{\text{gen}} \|} & \text{if } i \text{ has valid match} \\
0 & \text{otherwise}
\end{cases}
\end{equation}
The face matching reward is computed as $r_{\text{face}}(x, \mathcal{L}) = \frac{1}{N} \sum_{i=1}^N s_i$. In this case, count mismatches ($O \neq N$) result in zero similarity for unmatched references, naturally penalizing incorrect counts. By coupling identity evaluation with spatial proximity, this reward jointly optimizes spatial positioning and identity preservation, preventing failure modes discussed. Our results in Section~\ref{sec:quantitative} and Appendix E demonstrate its effectiveness over naive matching.

\noindent \textbf{Pose Rewards.}
To further reduce copy-paste artifacts and push for natural looking scenes, we compute a frontal pose reward, with the prompt ``Everyone is looking at the camera". From geometric head pose estimation methods~\cite{wu2017simultaneous}, we use facial landmarks to compute a score $\delta_i$ based on roll angle and facial symmetry for each detected face $i$. The pose reward is formulated as $r_{\text{pose}}(x, \mathcal{L}) = \frac{1}{O}\sum_{i=1}^O \delta_i$. We provide the formulation for $\delta_i$ in Appendix C. Additionally, we also explore fine-grained body pose control paired with with pose canvases, detailed in Appendix C.


\begin{figure}[t]
\centering
\includegraphics[width=\columnwidth]{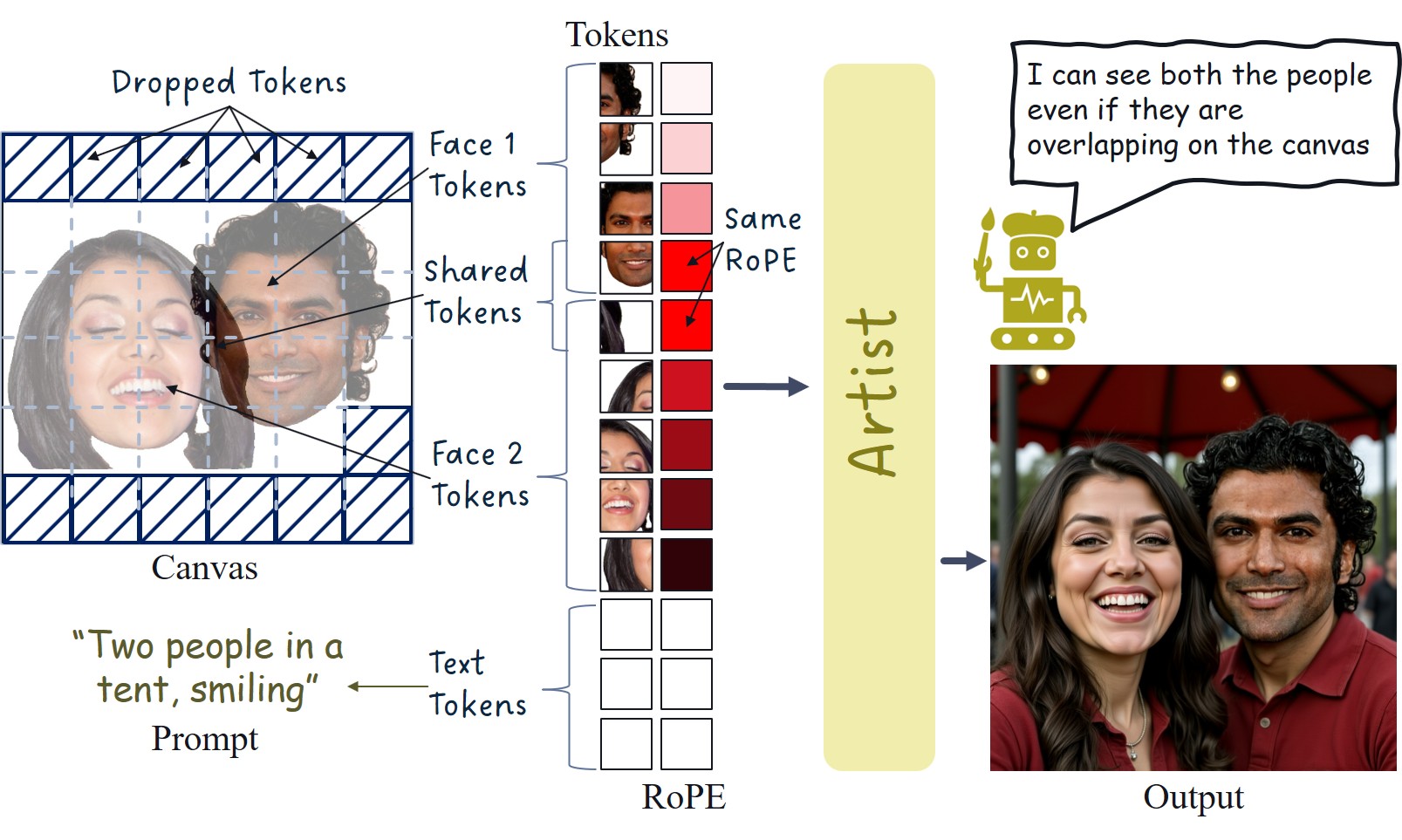}
\vspace{-2em}
\caption{We drop non-informative canvas tokens and assign identical positional encodings to overlapping regions, enabling the model to learn natural occlusion and depth ordering.}
\vspace{-1em}
\label{fig:token_sharing}
\end{figure}

\subsection{Dropping and Sharing Tokens}
\label{sec:token_sharing}

Feeding full canvas images at high resolution is computationally expensive and wastes computation on empty regions. Hence, we propose an efficient representation that preserves natural spatial relationships and enables the model to learn proper occlusion handling, see Figure~\ref{fig:token_sharing}.

We extract only tokens within the required spatial regions $\{b_1, \ldots, b_N\}$, reducing token counts by 2x on average (Fig.~\ref{fig:quantitative_analysis}a.). Additionally, when faces overlap ($b_i \cap b_j \neq \emptyset$), we feed both regions to the Kontext model but assign \textit{identical} RoPE positional embeddings to them: $\text{RoPE}(\text{tokens}(b_i)) = \text{RoPE}(\text{tokens}(b_j)) = \text{RoPE}(b_i \cup b_j)$. This signals spatial competition, forcing the model to resolve conflicts through strategies such as depth layering, spatial rearrangement, or partial occlusions, rather than pasting both identities at the same location. See Appendix E for visual results on effectiveness of token sharing.

\subsection{Training Curriculum}
\label{sec:training}
We employ a curriculum learning strategy over person count to stabilize early training. We partition training data into buckets based on number of faces per sample, $N \in \{2, 3, 4, 5, 6, 7\}$. For the first $\tau$ epochs, we sample equally from 2- and 3-person scenes. After $\tau$ epochs, we sample uniformly across all buckets. The transition probability is:
\begin{equation}
p(N|t) = \begin{cases}
\frac{1}{2} & \text{if } N \in \{2, 3\} \text{ and } t \leq \tau \\
\frac{1}{6} & \text{otherwise (for } t > \tau \text{)}
\end{cases}
\end{equation}
This curriculum is motivated by the base model's (Flux-Kontext) tendency to generate 1-2 faces reliably while failing at 3+ faces. Starting with simpler scenes prevents early training collapse and allows the model to first establish stable identity-preserving behavior before encountering harder compositional scenarios. Gradually introducing larger person counts then enables the model to generalize its learned spatial and identity reasoning to more complex scenes without catastrophic forgetting of simpler cases.

%% file: sec/4_results.tex
\begin{figure*}[t]
\centering
\includegraphics[width=\textwidth]{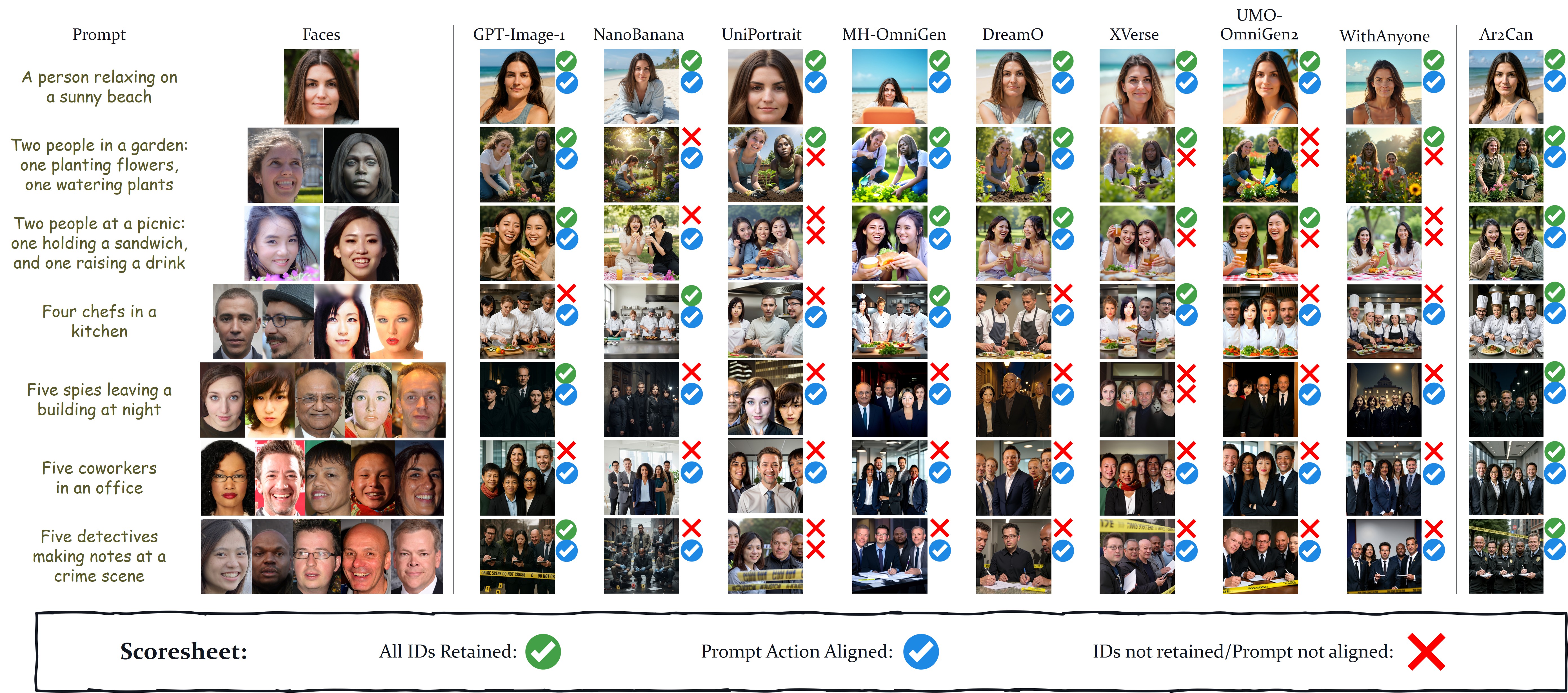}
\caption{\textbf{Qualitative comparison with state-of-the-art methods.} Left: reference images and text prompts from MultiHuman-Testbench. Right: outputs from all methods. The scorecards indicate ID preservation and prompt alignment. State-of-the-art methods frequently fail at either identity preservation or prompt alignment, while Ar2Can consistently achieves both across diverse multi-human scenes.}
\label{fig:results_comparison}
\end{figure*}

\section{Experiments}
\label{sec:experiments}

\subsection{Experimental Setup}
\label{sec:setup}

\noindent \textbf{Datasets.}
We evaluate on two benchmarks: \textbf{MultiHuman-Testbench (MHTB)}~\cite{borse2025multihuman} with 1,800 samples and 5550 faces, uniformly spanning 1-5 people per sample; and \textbf{Multi-ID-Test}~\cite{xu2025withanyone} with 1-4 faces biased toward 1-2 person scenes.

\noindent \textbf{Evaluation Metrics.}
Following the setup in MHTB~\cite{borse2025multihuman}, we employ four complementary metrics: \textbf{Count Accuracy} for correct number of generated people; \textbf{Hungarian ID Similarity} evaluates identity preservation using facial similarity with Hungarian matching to establish and quantify optimal face correspondences; \textbf{HPSv2} measures overall image quality and prompt alignment; and \textbf{Action-S/C} probes action correctness separately for simple and complex actions, using Gemini Flash as the MLLM. We also report the \textbf{Unified Metric}~\cite{borse2025multihuman} which balances identity preservation and prompt alignment.

\noindent \textbf{Implementation Details.}
\textbf{Architect-A} uses Qwen-2.5-0.5B-Instruct, trained for 20K steps. \textbf{Architect-B} fine-tunes Flux-Schnell for 240 GRPO epochs. Our \textbf{Artist} is a trained Flux-Kontext for 300 GRPO epochs. We use curriculum scheduling over person counts, annealing from $\{2,3\}$ to uniform $\{2,\ldots,7\}$ over $\tau=100$ epochs. All additional details and hyperparameters are in the Appendix B.


\begin{table*}[htbp]
    \renewcommand{\arraystretch}{1.5}
    \setlength{\tabcolsep}{7pt} 
    \fontsize{7.5pt}{6.75pt}\selectfont
    \centering
    \begin{tabular}{l|l|cccccc}
        \hline
         & \multirow{2}{*}{Model} & \multicolumn{6}{c}{Metrics} \\
         & & Count & Multi-ID & HPS & Action-S & Action-C & Unified \\
        \hline
        \multicolumn{8}{c}{\cellcolor{lightyellow}\textbf{MultiHuman-TestBench}} \\
        \hline
        \cellcolor{lightpurple} 
            & GPT-Image-1 & 87.9 & 28.8 & \cellcolor{green!10}30.3 & \cellcolor{green!10}97.0 & \cellcolor{green!10}91.1 & 55.8 \\
        \cellcolor{lightpurple} \multirow{-2}{*}{Proprietary}
            & Nanobanana  & 75.0 & \cellcolor{red!10}20.6 & 30.1 & \cellcolor{green!20}98.9 & \cellcolor{green!20}95.5 & 44.9 \\
        \hline
        \cellcolor{lightblue} 
            & UniPortrait   & \cellcolor{red!10}58.5 & 44.2 & 28.9 & 76.2 & 67.2 & 51.7 \\
        \cellcolor{lightblue} 
            & OmniGen       & 60.5 & 49.4 & 26.2 & 87.5 & 71.3 & 59.2 \\
        \cellcolor{lightblue} 
            & MH-OmniGen    & 60.3 & 54.5 & 26.3 & 91.6 & 72.9 & 61.6 \\
        \cellcolor{lightblue} 
            & DreamO       & 61.2 & 34.7 & 28.5 & 86.2 & 81.5 & 59.7 \\
        \cellcolor{lightblue} 
            & UMO-UNO       & \cellcolor{red!20}41.6 & 23.1 & \cellcolor{red!20}19.8 & \cellcolor{red!20}48.8 & \cellcolor{red!20}45.0 & \cellcolor{red!20}33.2 \\
        \cellcolor{lightblue} 
            & UMO-OmniGen2  & 70.5 & 46.4 & 29.3 & 83.6 & 80.1 & 60.4 \\
        \cellcolor{lightblue} \multirow{-7}{*}{\parbox{1.7cm}{Open-Source}}
            & XVerse       & 81.7 & 30.6 & \cellcolor{red!10}25.5 & \cellcolor{red!10}66.2 & \cellcolor{red!10}61.3 & 52.7 \\
        \hline
        \cellcolor{lightblue!60}
            & ID-Patch (Arch-B)    & 86.3 & 55.1 & 26.8 & 85.4 & 73.2 & 63.3 \\
           \cellcolor{lightblue!60} & WithAnyone (Arch-B)    & \cellcolor{green!10}89.8 & 44.3 & 29.7 & 82.5 & 77.0 & 62.6 \\
        \cellcolor{lightblue!60} \multirow{-3}{*}{\parbox{1.7cm}{Our Architect}}
            & Kontext (Arch-B) & 80.7 & \cellcolor{red!20}14.5 & 29.2 & 83.0 & 78.3 & \cellcolor{red!10}38.2 \\
        \hline
        \cellcolor{lightgreen} 
            & Ar2Can (Arch-B)    & 86.9 & \cellcolor{green!20}68.2 & \cellcolor{green!20}30.8 & 86.2 & 82.0 & \cellcolor{green!20}72.4 \\
        \cellcolor{lightgreen} \multirow{-2}{*}{\parbox{1.7cm}{Ours}}
            & Ar2Can (Arch-A)    & \cellcolor{green!20}90.2 & \cellcolor{green!10}67.6 & 30.2 & 86.3 & 77.6 & \cellcolor{green!10}72.2 \\
        \bottomrule
    \end{tabular}
    \caption{Comparison with state-of-the-art methods on reference-based multi-human generation. Color coding: \colorbox{green!20}{highest} and \colorbox{red!20}{lowest}.} 
    \label{tab:multihuman_recent}
    \vspace{-5pt}
\end{table*}

\begin{table}[htbp]
    \centering
    \resizebox{\linewidth}{!}{
    \begin{tabular}{l|ccc}
        \hline
        \multirow{2}{*}{Model} & \multicolumn{3}{c}{Metrics} \\
         & Multi-ID (Ref) & Multi-ID (GT) & Prompt Align \\
        \hline
        \multicolumn{4}{c}{\cellcolor{lightyellow}\textbf{MultiID-Test}} \\
        \hline
        OmniGen2 & \cellcolor{red!20}28.8 & \cellcolor{red!20}22.2 & \cellcolor{green!20}30.5 \\
        DreamO & \cellcolor{red!10}39.6 & \cellcolor{red!10}27.2 & \cellcolor{red!10}29.8 \\
        UMO-OmniGen2 & 47.5 & 32.0 & 30.0 \\
        WithAnyone & \cellcolor{green!10}50.1 & \cellcolor{green!10}33.8 & \cellcolor{red!20}28.6 \\
        \hline
        \textbf{Ours (Ar2Can)} & \cellcolor{green!20}54.3 & \cellcolor{green!20}36.5 & \cellcolor{green!10}30.1 \\
        \bottomrule
    \end{tabular}}
    \caption{Evaluation on MultiID-2M test set. ID Sim (Input): similarity to reference images, ID Sim (GT): similarity to ground truth identities, and Prompt Align assesses text-image alignment.}
    \label{tab:multiid2m}
    \vspace{-2em}
\end{table}




\subsection{Results}
\label{sec:quantitative}

Table~\ref{tab:multihuman_recent} and~\ref{tab:multiid2m} compare Ar2Can to state-of-the-art methods on MultiHuman-Testbench~\cite{borse2025multihuman} and Multi-ID-Test respectively. We evaluate proprietary systems (GPT-Image-1, Nanobanana), open-source methods (UniPortrait~\cite{he2025uniportrait}, OmniGen~\cite{xiao2025omnigen, wu2025omnigen2}, DreamO~\cite{mou2025dreamo}, UMO~\cite{cheng2025umo}, XVerse~\cite{chen2025xverse}, WithAnyone~\cite{xu2025withanyone}), and our proposed method Ar2Can.

\noindent \textbf{MultiHuman-Testbench Results.} 
Ar2Can with Architect-A achieves the highest count accuracy (\textbf{90.2}), significantly outperforming all baselines including WithAnyone (89.8) and GPT-Image-1 (87.9). This demonstrates the effectiveness of our LLM-based autoregressive layout generation for accurate person counting. For identity preservation, Ar2Can with Architect-B achieves the best Multi-ID score (\textbf{68.2}), substantially exceeding the previous best open-source method MH-OmniGen (54.5) by \textbf{13.7 points}. This validates our spatially-grounded face matching reward and Hungarian correspondence. Ar2Can maintains the highest image quality, with Architect-B achieving HPS of \textbf{30.8}. On the unified metric that balances all objectives, Ar2Can achieves \textbf{72.4} and \textbf{72.2} for Architect-B and A respectively, substantially outperforming all baselines. Observing the \textbf{proprietary methods}, GPT-Image and Nanobanana show strong scores in Action generation, but they significantly underperform on identity preservation. This is likely due to feature hallucination. Ar2Can heavily surpasses these methods on Multi-ID similarity while maintaining competitive Prompt Alignment and Quality score. We also demonstrate comparable Action scores with open-source methods.

\begin{figure*}[t]
\centering
\includegraphics[width=0.95\textwidth]{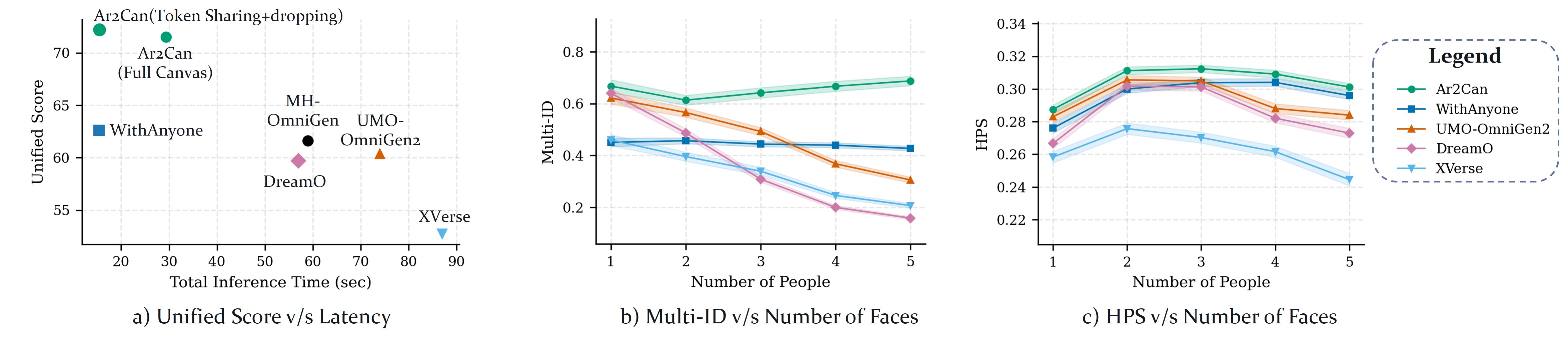}
\vspace{-1em}
\caption{\textbf{Quantitative analysis and scalability.} (a) Latency-quality trade-off on A100 GPU. (b) Multi-ID similarity vs. person count. (c) HPS vs. person count.  Ar2Can maintains consistent identity preservation and Prompt Alignment across 1-5 people.}
\vspace{-1em}
\label{fig:quantitative_analysis}
\end{figure*}

\noindent \textbf{Multi-ID-Test Results.}
Table~\ref{tab:multiid2m} shows results on Multi-ID-Test, which is heavily biased toward 1-2 person scenes where the multi-human challenge is less pronounced. Despite this bias, Ar2Can achieves the best performance with Multi-ID (Ref) of \textbf{54.3}, outperforming WithAnyone (50.1) by 4.2 points. Notably, Multi-ID (GT) measures similarity to a different view of the ground truth identity, testing view-invariant identity preservation. Ar2Can achieves \textbf{36.5}, exceeding WithAnyone (33.8) by 2.7 points, while maintaining competitive prompt alignment (\textbf{30.1}). The smaller performance gaps compared to MHTB reflect the dataset's bias toward simpler 1-2 person scenarios, where baseline methods face fewer identity conflicts.

\noindent \textbf{Architect Trade-offs.} 
Architect-A (Qwen-based) excels at count accuracy (90.2 vs 86.9) due to stronger language understanding, while Architect-B (Flux-based) achieves superior Action scores thanks to better spatial priors from its 2D nature. Both substantially outperform prior work on the Unified metric, demonstrating balanced performance across all evaluation dimensions.

\noindent \textbf{Qualitative:} Figure~\ref{fig:results_comparison} presents a comprehensive qualitative comparison of Ar2Can(Arch-B) against state-of-the-art methods on MultiHuman-Testbench samples. We evaluate each generation using two criteria: ID preservation, indicating whether faces match reference identities, and prompt alignment, assessing action correctness. As shown in the scorecards, existing methods exhibit a clear trade-off between these objectives. Methods optimized for identity preservation (e.g., XVerse, UniPortrait) often fail to execute complex actions correctly, while those achieving better prompt alignment (e.g., GPT-Image-1, Nanobanana) suffer from identity hallucination or blending artifacts. In contrast, Ar2Can consistently achieves both objectives across simple and complex action scenarios. The two-stage architecture's explicit spatial grounding prevents identity merging, while our Hungarian-based face matching reward ensures correct identity-location correspondence. Additional qualitative results and failure cases are provided in the Appendix E.

\noindent\textbf{Latency Analysis:} Figure~\ref{fig:quantitative_analysis}(a) shows the latency-quality trade-off on A100 GPU ($1024 \times 1024$, 3 identities). As observed, Ar2Can obtains a heavily favorable trade-off, and sharing/dropping tokens significantly helps boost it further.



\begin{table}[htbp]
    \vspace{-0.5em}
    \centering
    \resizebox{\linewidth}{!}{
    \begin{tabular}{l|ccc|ccc}
        \hline
        \multirow{2}{*}{\centering Model} & \multicolumn{3}{c|}{Components} & \multicolumn{3}{c}{Metrics} \\
        \cline{2-7}
          & \makecell{Simple\\Matching}
          & \makecell{Hungarian\\Centroid\\Matching}
          & \makecell{Curriculum\\Training}
          & \makecell{Count}
          & \makecell{Multi-ID}
          & \makecell{HPS} \\
        \hline
        Baseline (Kontext) & & & & \cellcolor{green!10}80.7 & 14.5 & 29.2 \\
        \hline
        & \checkmark & & & 75.6 & 55.2 & 27.6 \\
        & & \checkmark & & 80.1 & \cellcolor{green!10}60.3 & \cellcolor{green!20}30.9 \\
        \textbf{Ours (Full)} &  & \checkmark & \checkmark & \cellcolor{green!20}86.9 & \cellcolor{green!20}68.2 & \cellcolor{green!10}30.8 \\
        \bottomrule
    \end{tabular}}
    \vspace{-0.5em}
    \caption{Ablation study on key training components. We progressively add each component to measure its contribution to count accuracy, identity preservation, and image quality.}
    \label{tab:ablation}
    \vspace{-1em}
\end{table}

\noindent\textbf{Ablation Study:} Table~\ref{tab:ablation} analyzes contributions from two of our proposed components, the hungarian centroid matching reward and our proposed training curriculum. Simple matching (extracting faces at exact locations) improves Multi-ID but degrades both count (80.7→75.6) and HPS (29.2→27.6) due to copy-paste artifacts and unnatural face sizes. Hungarian centroid matching on the other hand, provides more flexible spatial constraints, recovering quality while improving Multi-ID. Curriculum training gradually introduces complexity from simpler 2-3 person scenes to larger counts, enabling the full model to achieve 86.9 count accuracy and 68.2 Multi-ID. Further analysis and visual comparisons are provided in Appendix D and E.

\noindent \textbf{User Preference Study.}
We conducted a user study comparing Ar2Can against DreamO~\cite{mou2025dreamo} and XVerse~\cite{chen2025xverse} on 25 triplet samples with 3-5 input identities and complex scene prompts. 25 evaluators rated each generation across three criteria: Prompt Alignment, Identity Preservation, and Overall Quality, selecting their preferred method per prompt. As shown in Table~\ref{tab:user_study}, Ar2Can is strongly preferred across all criteria, winning 88\% of prompts, confirming that our quantitative gains align with human perception.

\begin{table}[h]
    \vspace{-0.5em}
    \renewcommand{\arraystretch}{1.2}
    \fontsize{8.0pt}{9pt}\selectfont
    \centering
    \begin{tabular}{l|c|ccc}
        \hline
        Method & \%\,Wins$\uparrow$ & Prompt Algn.$\uparrow$ & ID Sim.$\uparrow$ & Quality$\uparrow$ \\
        \hline
        DreamO  & 4  & 19.5 & 8.3  & 9.8  \\
        XVerse  & 8  & 35.7 & 31.2 & 31.2 \\
        \rowcolor{gray!15}
        \textbf{Ar2Can} & \textbf{88} & \textbf{75.2} & \textbf{79.8} & \textbf{70.1} \\
        \hline
    \end{tabular}
    \vspace{-0.5em}
    \caption{Human preference study. 25 evaluators rated 25 triplet samples. \%\,Wins denotes the percentage of prompts for which a method was preferred overall.}
    \label{tab:user_study}
\end{table}

\vspace{-1.5em}
\section{Conclusion}
\label{sec:conclusion}
\vspace{-0.5em}

We presented Ar2Can, a two-stage framework that resolves identity preservation challenges in multi-human generation through explicit spatial decomposition. Our Architect generates layouts while the Artist renders photorealistic images. This design prevents the identity merging and swapping failures that plague end-to-end methods. By utilizing mostly synthetic data, and no real multi-human images, we demonstrate how Ar2Can achieves state-of-the-art performance on MultiHuman-Testbench. It substantially outperforms both proprietary and open-source baselines across all metrics. The modular architecture supports multiple Architect designs, efficient token sharing reduces computation, and our hybrid training approach eliminates the need for large-scale multi-human datasets. 


%% file: sec/X_suppl.tex
\onecolumn
\appendix
\appendixpage
\counterwithin{figure}{section}
\counterwithin{table}{section}
\renewcommand{\contentsname}{\Large\textbf{Contents}}
\setcounter{tocdepth}{3}
\vspace{1em}
\begin{center}
\rule{0.8\textwidth}{1pt}
\end{center}
\tableofcontents
\vspace{1em}
\begin{center}
\rule{0.8\textwidth}{1pt}
\end{center}
\vspace{2em}

\section{Introduction}
This appendix is comprehensive supplementary material to support our main paper. We organized this content into six sections that progressively detail our implementation, methodology, experimental results, and analysis.

Section~\ref{app:implementation} presents complete implementation details and hyperparameters for all components of Ar2Can, including both the Architect variants and our Artist module. This includes the final run training curves. Section~\ref{app:method} extends our method description with detailed algorithms, architectural specifications, and training procedures omitted from the main paper due to space constraints. Section~\ref{app:quant} provides extensive quantitative analysis. This includes any additional ablation studies, scaling experiments, and component-wide performance breakdowns. Section~\ref{app:qual} presents comprehensive qualitative results, visualizing failure modes, architectural comparisons, and the effectiveness of our proposed components. Finally, Section~\ref{app:limitations} discusses current limitations of our approach and outlines promising directions for future research. Together, these sections provide the complete technical details necessary for reproducing our results and understanding the full scope of our contributions.

\section{Implementation Details and Hyperparameters}
\label{app:implementation}

This section provides comprehensive implementation details for reproducing our results. We begin with our training dataset construction and prompt curation strategy (Section~\ref{app:implementation}), followed by model architectures and detection tools used throughout our pipeline. We then detail the complete training procedures for both Architect variants and the Artist, including hyperparameters, optimization schedules, and hardware configurations. Finally, we describe our inference setup and computational requirements.

\subsection{Training Dataset}
Our training dataset consists of 60,000 carefully curated prompts designed to capture diverse multi-human scenarios. Each prompt describes group scenes containing 2-7 people engaged in various activities and contexts. The captions were generated using GPT-5 to ensure high-quality, diverse descriptions that encompass a wide range of:

\begin{itemize}
    \item \textbf{Social contexts}: Family gatherings, business meetings, friend groups, professional teams, recreational activities
    \item \textbf{Settings}: Indoor and outdoor environments, formal and informal occasions, workplace and leisure contexts
    \item \textbf{Activities}: Collaborative tasks, social interactions, professional activities, recreational pursuits
    \item \textbf{Group compositions}: Varying numbers of individuals (2-7) with diverse demographic representations
\end{itemize}

for 30$\%$ of the prompts, we add the tag {\fontfamily{cmss}\selectfont \textcolor{darkyellow}{Everyone is looking at the camera}}. These prompts activate the frontal pose reward during training. The prompts were designed to avoid overlap with evaluation datasets while maintaining sufficient diversity to train robust multi-human generation capabilities. The following are 5 examples of these prompts.

\begin{itemize}
  \item {\fontfamily{cmss}\selectfont \textcolor{darkyellow}{Seven people on the desert dunes, hazy sun, diverse faces, clear faces visible, studio-quality, vivid detail}}
  \item {\fontfamily{cmss}\selectfont \textcolor{darkyellow}{Six people in an astronomy studio, Clean composition, Professional portrait, Portrait photography, Soft shadows, Natural lighting, Even exposure}}
  \item {\fontfamily{cmss}\selectfont \textcolor{darkyellow}{Three people in an aviation observatory, Sharp focus, Clean composition, Bokeh background, Color graded, Smiling expressions, Well lit}}
  \item {\fontfamily{cmss}\selectfont \textcolor{darkyellow}{Five people in a dawn-lit bakeshop, Studio quality, Even exposure, Group harmony, Cinematic lighting, Portrait photography, Soft shadows}}
  \item {\fontfamily{cmss}\selectfont \textcolor{darkyellow}{Seven people on a coastal boardwalk, afternoon light, diverse faces, clear faces visible, ultra-realistic, 8K resolution. Everyone is looking at the camera.}}
\end{itemize}

The canvas is then constructed for each prompt as discussed in Algorithm~\ref{alg:canvas_construction}, with the sampling probabilities $p_1=0.5, p_2=0.4, p_3=0.1$. We try to reduce the amount of synthetic faces, even though DisCo has been trained to generate diverse faces.

\subsection{Models}
\textbf{Architect-A} uses Qwen-2.5-0.5B. \textbf{Architect-B} uses Flux-Schnell (1B parameters, 4 steps). The \textbf{Artist} is a Flux-Kontext (12B parameters). We use RetinaFace~\cite{deng2020retinaface} (ResNet-50 backbone) for face detection with confidence threshold 0.5. To compute ArcFace embeddings~\cite{deng2019arcface}, we use the Anetlopev2\footnote{\url{https://github.com/deepinsight/insightface}} model, producing 512-dimensional features. For pose computation, we use the rtmlib library~\footnote{\url{https://github.com/Tau-J/rtmlib}} with their default model.

\subsection{Training}

\begin{figure}[h]
  \centering
  \includegraphics[width=0.9\linewidth]{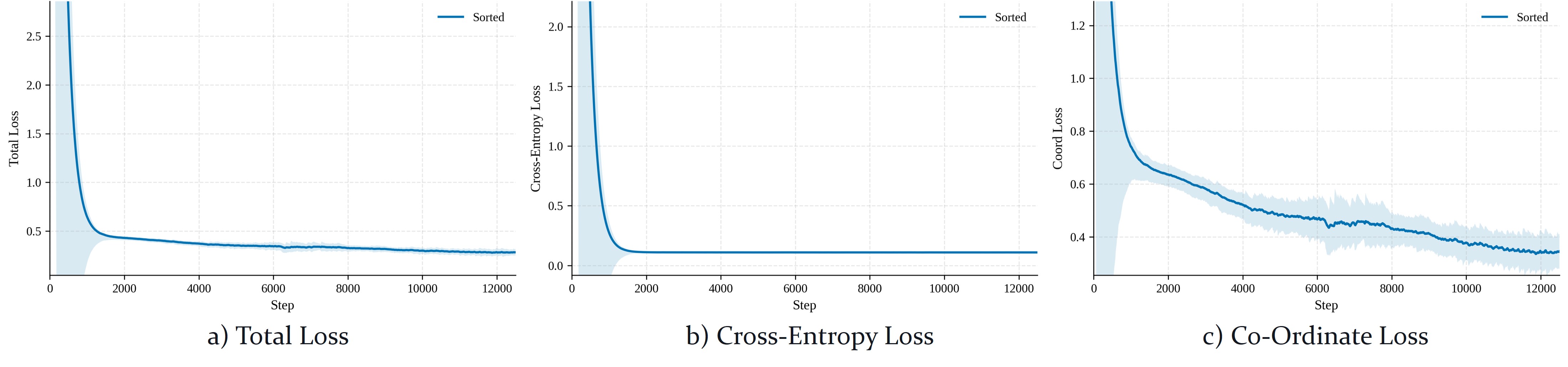}
  \caption{\textbf{Training curves of Architect-A (LLM-based).}
The cross-entropy loss converges quickly, indicating that the model stabilizes its estimation of the number of bounding boxes after roughly 2000 steps. After this point, training is dominated by the optimization of bounding-box coordinate regression.}
  \label{fig:arch_a_curves}
\end{figure}

For \textbf{Architect-A}, we fine-tune Qwen for 20K steps using AdamW with a learning rate of $10^{-5}$ and a batch size of 128. We begin with a 100-step warm-up phase in which the learning rate increases linearly from zero to its maximum value, followed by a cosine decay schedule that gradually reduces it back to zero. We set $\lambda_{\text{coord}} = 0.8$ to balance the coordinate regression loss and the cross-entropy loss. All coordinates are normalized to the range $[0, 1]$. The coordinate regression head $f_{value}$ is a 4-layer MLP. For coordinate embedding $f_\text{embed}$, we first map the normalized coordinates back to the original $[0, 1024]$ range, convert them to integer values, and feed them into a discrete embedding layer. These embeddings are added to the token embeddings to provide positional information and allow the model to distinguish different coordinate tokens. The streams of $f_{\text{value}}$ and $f_{\text{embed}}$ are activated only when $f_{\text{token}}$ predicts that the next token corresponds to a coordinate token \texttt{<C>}. Architect-A training takes approximately 6 GPU-hours on a single A100-80GB.

Figure~\ref{fig:arch_a_curves} presents the SFT training progression of Architecture-A. 
The model adapts rapidly to the counting task, with the cross-entropy loss converging within the first 2000 steps. After this point, the remaining optimization primarily focuses on refining bounding-box coordinate regression. Although the bounding-box count is an important component of the task, the learning of coordinate distributions plays a central role in shaping the model's spatial understanding. The convergence behavior of the coordinate loss reflects how effectively the model learns the geometric patterns in the data.

\begin{figure}[h]
  \centering
  \includegraphics[width=0.6\linewidth]{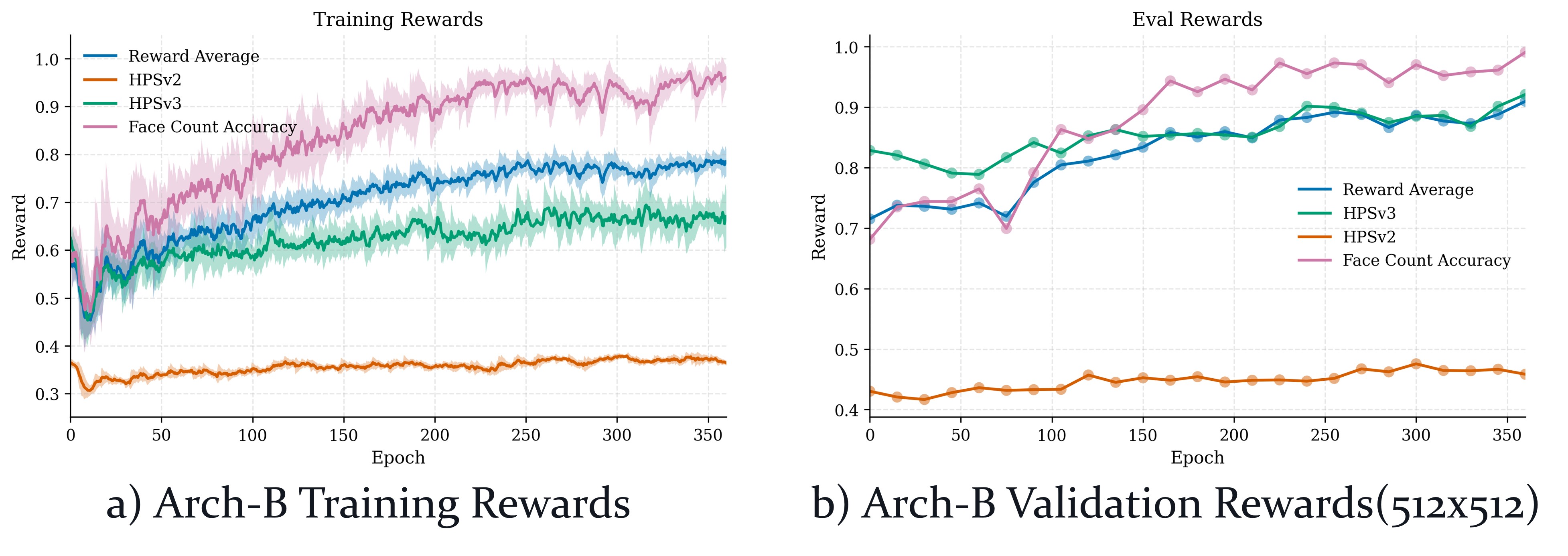}
  \caption{\textbf{Architect-B training and validation reward curves.} 
We observe steady improvement in both count accuracy and prompt alignment/aesthetic quality (HPS) rewards during training and evaluation, with count accuracy stabilizing earlier than HPS scores.}
  \label{fig:arch_b_curves}
\end{figure}

\paragraph{Architect-B Training.} We implement Architect-B using the public \texttt{flow\_grpo}\footnote{\url{https://github.com/yifan123/flow_grpo}} framework with Flux-Schnell pipeline, training in bf16 mixed precision on 512×512 images. To save on training memory, we train a LoRA with Rank=64 instead of training the full network. Training uses 3 timesteps for reward computation and evaluation. We train for 240 epochs with batch sizes of 3 (train) and 16 (test), with a group size of 21. The compositional reward function combines count accuracy ($\alpha = 0.5$) and prompt alignment via HPS. We use both HPSv2 and HPSv3 ($\beta = 0.25$ each), with KL regularization weight $\beta_{\text{KL}} = 0.01$ to stabilize learning. Training is distributed across 8$\times$H100-80GB GPUs on a single node, of which 1 dedicated GPU is used for HPSv3 reward computation and 7 GPUs for training. We use a learning rate of $1 \times 10^{-4}$ with EMA enabled. Face detection for count accuracy uses blob analysis on the 3-step generated images. Total training time to 240 epochs is approximately \textbf{140 GPU-hours}.

Figure~\ref{fig:arch_b_curves} demonstrates the training progression of Architect-B across both reward components throughout the GRPO fine-tuning process. The curves show consistent improvement in count accuracy and prompt alignment (HPSv3) metrics during both training and evaluation phases. Count accuracy shows rapid improvement in the first 200 epochs, stabilizing around epoch 240, similar to HPSv3 scores. The model achieves strong performance on the test set by epoch 240, with minimal overfitting observed. Please note that these scores are on the val set(different from our testsets in the paper), and also for model inference at 512x512, at 18 timestep inference; hence, the numbers are different from scores in the paper.

\begin{figure}[h]
  \centering
  \includegraphics[width=0.6\linewidth]{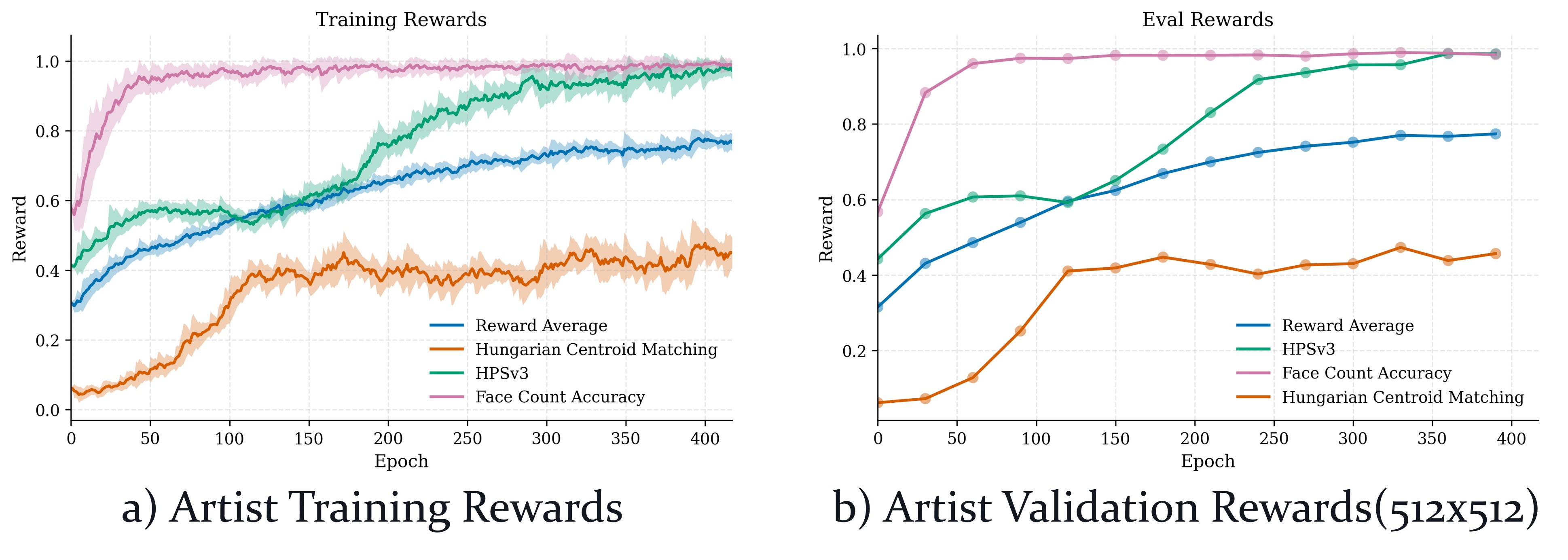}
  \caption{\textbf{Artist training and evaluation reward curves.} 
We observe steady improvement in all compositional rewards (count accuracy, HPSv3, face matching, and pose alignment) during training and evaluation. The curriculum transition at epoch 100 is visible as a brief perturbation in HPSv3, after which training stabilizes with continued improvement.}
  \label{fig:artist_curves}
\end{figure}

\paragraph{Artist Training.} We implement the Artist using the public \texttt{flow\_grpo} framework with Flux-Kontext pipeline, training in bf16 mixed precision on 512×512 images. To save on training memory, we train a LoRA with Rank=64 instead of training the full network. Training uses 18 timesteps for reward computation and evaluation, with classifier-free guidance of 2. We train for 300 epochs with batch sizes of 3 (train) and 16 (test), with a group size of 21. The compositional reward function combines count accuracy ($\alpha = 0.2$), image quality via HPSv3 ($\beta = 0.4$), hungarian centroid face matching ($\zeta = 0.3$), and frontal pose alignment ($\eta = 0.1$) components, with KL regularization weight $\beta_{\text{KL}} = 0.00$ to stabilize learning. We apply curriculum learning with transition epoch $\tau = 100$, sampling equally from 2-3 person scenes for the first 100 epochs, then uniformly from 2-7 person scenes. Training is distributed across 24 H100-80GB GPUs on 3 nodes, consisting of 3 dedicated GPUs for HPSv3 reward servers (1 per node). Hence, we use 21 GPUs for training (7 per node) and 3 for the reward servers. We use a learning rate of $1 \times 10^{-4}$ with EMA enabled. Total training time to 300 epochs is approximately \textbf{1000 GPU-hours}, and on our setup it takes \textbf{2 days}. 

Figure~\ref{fig:artist_curves} demonstrates the training progression of the Artist across all four compositional reward components throughout the GRPO optimization process. The curves show consistent improvement in count accuracy, image quality (HPSv3), spatially-grounded face matching(including frontal pose alignment) during both training and evaluation phases. Face matching rewards exhibit the most dramatic improvement, particularly in the first 150 epochs, reflecting the model's learning to preserve identities while maintaining spatial correspondence. Count accuracy stabilize relatively quickly, while HPSv3 continues gradual improvement throughout training. The curriculum transition at epoch 100 (from 2-3 person scenes to all counts) is visible as a brief perturbation in the HPSv3 curve, after which training continues smoothly. Please note that the evaluation scores are on the val set(different from our testsets in the paper), and also for model inference at 512x512, at 18 timestep inference; hence, the numbers are different from scores in the paper.

\subsection{Inference}
At inference, we sample an Architect layout and generate the final image. For Architect-A, layout generation is deterministic (greedy decoding). For Architect-B, we use 4-step flow sampling with 0 guidance scale. The Artist uses 28-step generation with guidance scale of 2. For the Artist, we apply the token saving and pass input faces through separate canvases. In overlapping tokens, we pass the same RoPE positional ID but pass both the tokens. All evaluations are at $1024 \times 1024$ resolution.

\begin{algorithm}[t]
\caption{Canvas Construction Pipeline}
\label{alg:canvas_construction}
\begin{algorithmic}[1]
\Require Data sources: multi-view $\mathcal{D}_1$, single-view $\mathcal{D}_2$, synthetic $\mathcal{D}_3$ with probabilities $p_1, p_2, p_3$
\Require Pose mode flag: \texttt{use\_pose}
\Ensure Training sample: canvas $C$, reference faces $\{I_{\text{ref},i}\}$, target image $I_{\text{DisCo}}$

\State \textbf{Stage 1: Scene Generation}
\State Sample target count $n \sim \text{Uniform}(2, 7)$
\State Generate $I_{\text{DisCo}} \sim \text{Flux-DisCo}(n)$ \Comment{Multi-person scene}

\State \textbf{Stage 2: Face Localization}
\State Detect face bounding boxes $\{b_1^{\text{DisCo}}, \ldots, b_n^{\text{DisCo}}\} \gets \text{RetinaFace}(I_{\text{DisCo}})$
\If{\texttt{use\_pose}}
    \State Estimate human poses $\{p_1^{\text{DisCo}}, \ldots, p_n^{\text{DisCo}}\} \gets \text{PoseEstimator}(I_{\text{DisCo}})$
\EndIf

\State \textbf{Stage 3: Reference Selection \& Augmentation}
\For{each face location $i = 1, \ldots, n$}
    \State Sample source $k \sim \text{Categorical}(p_1, p_2, p_3)$
    \State Sample reference face $I_{\text{ref},i}$ from $\mathcal{D}_k$
    \If{$k = 1$} \Comment{Multi-view}
        \State Use secondary view directly
    \ElsIf{$k = 2$ or $k = 3$} \Comment{Single-view or synthetic}
        \State Generate secondary view via:
        \State \hspace{1em} PuLID~\cite{guo2024pulid} (pose/expression variation), or
        \State \hspace{1em} Augmentations (rotation $\pm 15^\circ$, flip, brightness)
    \EndIf
\EndFor

\State \textbf{Stage 4: Canvas Composition}
\State Initialize blank canvas $C$ with resolution matching $I_{\text{DisCo}}$
\For{each face $i = 1, \ldots, n$}
    \State Paste $I_{\text{ref},i}$ onto $C$ at location $b_i^{\text{DisCo}}$
\EndFor
\If{\texttt{use\_pose}}
    \For{each person $i = 1, \ldots, n$}
        \State Overlay pose skeleton $p_i^{\text{DisCo}}$ onto $C$
    \EndFor
\EndIf

\State \Return $C$, $\{I_{\text{ref},i}\}_{i=1}^n$, $I_{\text{DisCo}}$
\end{algorithmic}
\end{algorithm}

\paragraph{Baselines.} 
We evaluate all baseline methods using their official open-source implementations with default parameters unless otherwise noted. For all the baseline methods, evaluations are at $1024 \times 1024$ resolution.

\noindent\textbf{UMO-UNO}~\cite{cheng2025umo}: We obtained the code from the official repository\footnote{\url{https://github.com/bytedance/UMO}} and used the default configuration with 25 inference steps and default guidance scale as recommended by the authors.

\noindent\textbf{X-Verse}~\cite{chen2025xverse}: We obtained the code from the official repository\footnote{\url{https://github.com/bytedance/XVerse}} and followed the default settings with 28 inference steps and default guidance scale.

\noindent\textbf{UMO-OmniGen2}~\cite{cheng2025umo}: We obtained the code from the official repository\footnote{\url{https://github.com/VectorSpaceLab/OmniGen}} and used the recommended configuration with 50 inference steps and default guidance scale.

\noindent\textbf{Dream-O}~\cite{mou2025dreamo}: We obtained the code from the official repository\footnote{\url{https://github.com/bytedance/DreamO}} and adopted the recommended Flux.1-turbo variant with 12 inference steps and default guidance scale as specified in the documentation.

\noindent\textbf{ID-Patch}~\cite{zhang2025id}: We obtained the code from the official repository\footnote{\url{https://github.com/bytedance/ID-Patch}} and used all default parameters from the official implementation without modification.

\noindent\textbf{WithAnyone}~\cite{xu2025withanyone}: We obtained the code from the official repository\footnote{\url{https://github.com/Doby-Xu/WithAnyone}} and used the default parameters except for the SigLIP weight, which we set to 0.5 as we found it to be the optimal trade-off between identity preservation and prompt alignment.

\section{Extended Method Section}
\label{app:method}

This section expands upon the methodological components described in the main paper with additional algorithmic details and formulations. We provide the complete canvas construction pipeline (Algorithm~\ref{alg:canvas_construction}) that generates our hybrid training data, followed by detailed formulations for our frontal pose reward and our optional body pose controlled Artist variant. We then present step-by-step training algorithms for all three components: Architect-A (supervised fine-tuning), Architect-B (GRPO with count and quality rewards), and the Artist (GRPO with compositional rewards including our novel Hungarian centroid matching). These algorithms complement the high-level descriptions in the main paper and provide the precise implementation details needed for replication.

\subsection{Canvas Construction Pipeline(Training)}

\begin{figure}[h]
    \centering
    \includegraphics[width=0.7\linewidth]{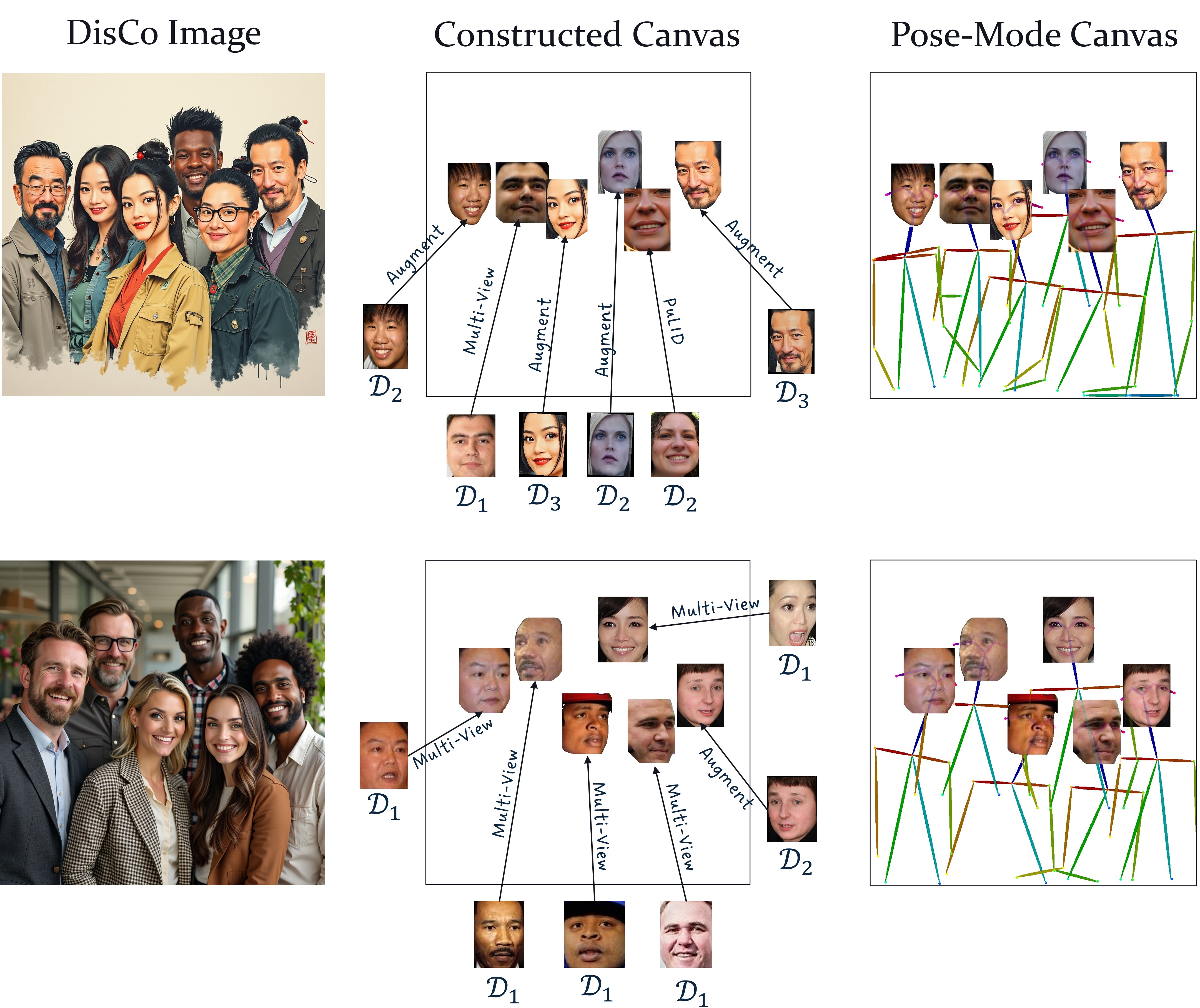}
    \caption{Illustration of the Canvas Construction Pipeline. \textbf{Left}: Image generated by DisCo~\cite{disco2025}, using the input prompt. \textbf{Middle}: Standard canvas. Each canvas combines reference faces sampled from different data sources: $\mathcal{D}_1$ (multi-view references with secondary views), $\mathcal{D}_2$ (single-view references with PuLID-generated or augmented secondary views), and $\mathcal{D}_3$ (the original synthetic face from DisCo). \textbf{Right}: If we want to perform pose-mode training, we overlay the human pose skeletons estimated from the DisCo images. We make sure to blend with lesser intensity in face bbox regions.}
    \label{fig:canvas_construction}
\end{figure}

We construct each training sample via a four-stage pipeline (as expanded in Algorithm~\ref{alg:canvas_construction}). This produces hybrid samples combining synthetic multi-person scenes with real reference faces. Figure~\ref{fig:canvas_construction} illustrates this process with example canvas samples showing reference faces drawn from different data sources ($\mathcal{D}_1$, $\mathcal{D}_2$, $\mathcal{D}_3$).

We start the process by generating a DisCo~\cite{disco2025} scene with $N$ people and detecting face locations (Stage 1-2). For generating the canvas in a pose-controlled training, we additionally estimate human pose skeletons from the DisCo image. Then, we initialize a blank canvas and replace each synthetic face with a reference face from our curated sources (Stage 3-4). In pose mode, we overlay the estimated pose skeletons onto the canvas, providing additional spatial guidance for action and pose alignment.

Each training sample comprises three components: the constructed canvas (with reference faces and optionally pose overlays), the original reference face images, and the DisCo generated target image. As illustrated in Figure~\ref{fig:canvas_construction}, reference faces are sampled from different data sources. These are $\mathcal{D}_1$ (multi-view), $\mathcal{D}_2$ (single-view with augmentation), and $\mathcal{D}_3$ (synthetic). This enables diverse identity representation while maintaining authentic facial appearances.

An important consideration is that while we show a single canvas with all faces in it, the actual training sample has multiple canvases, each with a separate face. This is so that we can provide the shared RoPE encodings illustrated in~\ref{fig:token_sharing}.

\subsection{Frontal Pose Reward Formulation}

We design a lightweight 2D frontal pose reward using 5-point facial landmarks (left eye, right eye, nose, left mouth, right mouth) detected by RetinaFace~\cite{deng2020retinaface}. For each detected face $i$, we compute a frontality score $\delta_i$ combining roll angle and yaw asymmetry:

\noindent\textbf{Roll Angle:} We compute the roll angle from eye landmarks: $\theta_{\text{roll},i} = |\arctan\left(\frac{y_{\text{re}} - y_{\text{le}}}{x_{\text{re}} - x_{\text{le}}}\right)|$, where $(x_{\text{le}}, y_{\text{le}})$ and $(x_{\text{re}}, y_{\text{re}})$ are left and right eye coordinates.

\noindent\textbf{Yaw Asymmetry:} We compute left-right facial asymmetry as $\Delta_{\text{sym},i} = \left| \|{\bf p}_{\text{le}} - {\bf p}_{\text{nose}}\| - \|{\bf p}_{\text{re}} - {\bf p}_{\text{nose}}\| \right|$, where $\|{\bf p}_a - {\bf p}_b\|$ denotes Euclidean distance between landmarks.

The frontality score applies Gaussian penalties to both components:

\begin{equation}
\delta_i = \exp\left(-\frac{\theta_{\text{roll},i}^2}{2\sigma_{\text{roll}}^2}\right) \cdot \exp\left(-\frac{\Delta_{\text{sym},i}^2}{2\sigma_{\text{yaw}}^2}\right)
\end{equation}

where $\sigma_{\text{roll}} = 15.0$ degrees and $\sigma_{\text{yaw}} = 0.25$ (normalized by inter-ocular distance). We threshold $\delta_i$ to zero if $\delta_i < 0.9$ to create a sharp distinction between frontal and non-frontal faces. The frontal pose reward is: $r_{\text{pose}}(x, \mathcal{L}) = \frac{1}{O}\sum_{i=1}^O \delta_i$, where $O$ is the number of detected faces. This reward is activated when prompts contain ``Everyone is looking at the camera'' (30\% of training data). We empirically observe this formulation to be lightweight and effective for reducing copy-paste artifacts in our multi-human generation setting.

\subsection{Canvas Layout Construction(Inference)}
\label{app:inference_canvas}

At inference, the canvas layout $\mathcal{L}$ is constructed from the Architect-predicted bounding boxes and the user-provided reference images. For each reference image, we first apply a segmentation model to extract the face region, removing background elements before placement. The segmented face is then aligned to its corresponding Architect-predicted bounding box coordinates using a face detection model, which localizes facial landmarks to ensure accurate spatial correspondence between the reference face and its target location on the canvas. Each aligned, segmented face is then pasted onto a blank white canvas at the predicted box location, producing the final layout $\mathcal{L}$ that is fed to the Artist alongside the text prompt and reference images.

\subsection{Body-Pose Controlled Artist}
\label{app:body_pose}

\textbf{Note:} This section describes an optional pose-controlled Artist variant explored in our experiments (Section~\ref{app:qual}). The main Ar2Can model uses only the frontal face reward described above.

Our pose-controlled Artist variant conditions on full-body pose skeletons overlaid on the canvas (pose-mode in Figure~\ref{fig:canvas_construction}), providing explicit control over body poses and actions beyond just face locations.

We introduce a body pose reward based on keypoint matching. For each generated image, we detect human keypoints in both the reference pose layout and generated image using a pose estimation model (rtmlib).
Next, we compute \textbf{Object Keypoint Similarity (OKS)}\footnote{\url{https://cocodataset.org/\#keypoints-eval}} between each pair of reference and generated persons:

\begin{equation}
\text{OKS}_{j} = \frac{1}{|K_j|} \sum_{k \in K_j} \exp\!\left(-\frac{d_k^2}{2 a_j \kappa_k^2}\right)
\end{equation}

where $d_k$ is the Euclidean distance between matched keypoints (e.g., left elbow in reference vs. generated), $a_j$ normalizes by the person's area, $\kappa_k$ is a per-keypoint constant from the COCO evaluation protocol (e.g., $\kappa_k = 0.026$ for shoulders, $0.107$ for eyes), and $K_j$ contains number of visible keypoints.

Since the number of detected people may differ from the reference count, we use \textbf{Hungarian matching} to find the optimal one-to-one correspondence between reference and generated persons that maximizes total OKS. The body pose reward aggregates across matched pairs:

\begin{equation}
r_{\text{body-pose}}(x, \mathcal{L}) = \frac{1}{O} \sum_{j=1}^O \text{OKS}_j
\end{equation}

where $O$ is the number of matched pairs. This reward encourages the model to reproduce both overall scene layout and fine-grained limb positions.

\textbf{Training:} When training the pose-controlled Artist, this body pose reward replaces the frontal face reward. The pose-controlled variant achieves accurate pose alignment but exhibits increased copy-paste artifacts compared to the standard Artist (Figure~\ref{fig:posemode_results}), representing a trade-off between pose controllability and rendering naturalness.


\subsection{Training Algorithms}

This section provides concise training algorithms for each component of Ar2Can. Algorithm~\ref{alg:architect_a_training} describes the supervised fine-tuning procedure for Architect-A, Algorithm~\ref{alg:architect_b_training} details the GRPO-based training for Architect-B, Algorithm~\ref{alg:hungarian_matching} presents our Hungarian centroid matching procedure for spatially-grounded face matching, and Algorithm~\ref{alg:artist_training} presents the complete Artist training procedure with compositional rewards.

\begin{algorithm}[h]
\caption{Architect-A Training (Supervised Fine-tuning)}
\label{alg:architect_a_training}
\begin{algorithmic}[1]
\Require Pre-trained LLM $\theta_{\text{LLM}}$ (Qwen-2.5-0.5B), dataset $\mathcal{D} = \{(p_i, \mathcal{L}_i)\}$
\State Initialize heads $f_{\text{value}}, f_{\text{embed}}$; extend tokenizer with \texttt{<SoL>}, \texttt{<EoL>}, \texttt{<C>}
\For{each batch $\{(p_j, \mathcal{L}_j)\}$ in $\mathcal{D}$}
    \State Sort coordinates in $\mathcal{L}_j$ (left-to-right, top-to-bottom)
    \State Forward: $h = \theta_{\text{LLM}}(p_j)$; predict tokens $\hat{y} = f_{\text{token}}(h)$; coords $\hat{b} = f_{\text{value}}(h)$
    \State Compute: $\mathcal{L}_{\text{total}} = \mathcal{L}_{\text{CE}}(\hat{y}, y_j) + \lambda_{\text{coord}}[\mathcal{L}_{\text{gIoU}}(\hat{b}, b) + \|\hat{b} - b\|_1]$
    \State Update: $\theta_{\text{LLM}}, f_{\text{value}}, f_{\text{embed}} \leftarrow \text{Adam}(\nabla \mathcal{L}_{\text{total}})$
\EndFor
\State \Return Fine-tuned Architect-A
\end{algorithmic}
\end{algorithm}

\subsubsection{Architect-A Training}

Algorithm~\ref{alg:architect_a_training} presents the supervised fine-tuning procedure for our LLM-based Architect-A. As describd in Section~\ref{sec:architect_a} of the main text, the model is trained to autoregressively generate structured layouts with special tokens, while coordinate regression and embedding heads enable continuous bounding box prediction. 

\begin{algorithm}[h]
\caption{Architect-B Training (GRPO)}
\label{alg:architect_b_training}
\begin{algorithmic}[1]
\Require Pre-trained T2I $\pi_{\text{ref}}$ (Flux-Schnell), prompts $\mathcal{P}$, group size $M$
\State Initialize policy $\pi_\theta \leftarrow \pi_{\text{ref}}$
\For{each prompt $p$ in $\mathcal{P}$}
    \State Sample group: $\{x_1, \ldots, x_M\} \sim \pi_\theta(\cdot | p)$
    \State Compute rewards: $r_i = \alpha \cdot \mathbbm{1}[n_{\text{pred}}(x_i) = n_{\text{target}}] + \beta \cdot \text{HPSv3}(x_i, p)$
    \State Compute advantages: $A_i = (r_i - \mu_G) / (\sigma_G + \epsilon)$
    \State GRPO loss: $\mathcal{L} = \sum_{i=1}^M A_i \log \frac{\pi_\theta(x_i|p)}{\pi_{\text{ref}}(x_i|p)} - \beta_{\text{KL}} \text{KL}(\pi_\theta || \pi_{\text{ref}})$
    \State Update: $\theta \leftarrow \text{Adam}(\nabla \mathcal{L})$
\EndFor
\State \Return Fine-tuned Architect-B
\end{algorithmic}
\end{algorithm}

\subsubsection{Architect-B Training}

Algorithm~\ref{alg:architect_b_training} describes the GRPO procedure for Architect-B, fine-tuning the model to generate accurate person counts and plausible spatial layouts. This is an extension from the explanation in Section~\ref{sec:architect_b} of the main text.

\begin{algorithm}[h]
\caption{Hungarian Centroid Matching}
\label{alg:hungarian_matching}
\begin{algorithmic}[1]
\Require Image $x$, layout $\mathcal{L} = \{b_i^{\text{pred}}\}$ with centroids $\{c_i^{\text{pred}}\}$, refs $\{I_{\text{ref},i}\}$
\State Detect faces: $\{b_j^{\text{det}}\} \leftarrow \text{RetinaFace}(x)$; compute centroids $\{c_j^{\text{det}}\}$
\State Build cost matrix: $C_{ij} = \|c_i^{\text{pred}} - c_j^{\text{det}}\|_2$
\State Hungarian assignment: $\pi^* = \arg\min_\pi \sum_{i=1}^{\min(N,O)} C_{i,\pi(i)}$
\For{$i = 1$ to $N$}
    \If{$i$ has valid match}
        \State $e_i^{\text{ref}} = f_{\text{ArcFace}}(I_{\text{ref},i})$; $e_{\pi^*(i)}^{\text{gen}} = f_{\text{ArcFace}}(\text{crop}(x, b_{\pi^*(i)}^{\text{det}}))$
        \State $s_i = \frac{e_i^{\text{ref}} \cdot e_{\pi^*(i)}^{\text{gen}}}{\|e_i^{\text{ref}}\| \|e_{\pi^*(i)}^{\text{gen}}\|}$
    \Else
        \State $s_i = 0$
    \EndIf
\EndFor
\State \Return $r_{\text{face}} = \frac{1}{N} \sum_{i=1}^N s_i$
\end{algorithmic}
\end{algorithm}

\subsubsection{Hungarian Centroid Matching for Face Rewards}

Algorithm~\ref{alg:hungarian_matching} presents our Hungarian centroid matching procedure, which establishes spatial correspondence via centroid distances, then evaluates identity similarity on matched pairs. 

\begin{algorithm}[h]
\caption{Artist Training (GRPO)}
\label{alg:artist_training}
\begin{algorithmic}[1]
\Require Pre-trained $\pi_{\text{ref}}$ (Flux-Kontext), dataset $\mathcal{T}$, curriculum epoch $\tau$, group size $M$
\State Initialize $\pi_\theta \leftarrow \pi_{\text{ref}}$; partition dataset: $\mathcal{T}_N$ for $N \in \{2,\ldots,7\}$
\For{epoch $t = 1$ to $N_{\text{epochs}}$}
    \State Sample from $\mathcal{T}_2 \cup \mathcal{T}_3$ if $t \leq \tau$, else sample uniformly from all $\mathcal{T}_N$
    \For{each sample $(C, \{I_{\text{ref}}\}, \mathcal{L}, p, I_{\text{target}})$}
        \State Sample group: $\{x_1, \ldots, x_M\} \sim \pi_\theta(\cdot | p, \mathcal{L}, \{I_{\text{ref}}\})$
        \For{$i = 1$ to $M$}
            \State $r_{\text{count}}(x_i) = \mathbbm{1}[|\text{RetinaFace}(x_i)| = N]$
            \State $r_{\text{hps}}(x_i, p) = \text{HPSv3}(x_i, p)$
            \State $r_{\text{face}}(x_i, \mathcal{L}) = \text{HungarianFaceMatching}(x_i, \mathcal{L}, \{I_{\text{ref}}\})$ (Alg.~\ref{alg:hungarian_matching})
            \State $r_{\text{pose}}(x_i, \mathcal{L}) = \frac{1}{O}\sum_{j=1}^O \delta_j$ (frontality scores)
            \State $r_i = \alpha r_{\text{count}} + \beta r_{\text{hps}} + \zeta r_{\text{face}} + \eta r_{\text{pose}}$
        \EndFor
        \State Compute advantages: $A_i = (r_i - \mu_G) / (\sigma_G + \epsilon)$
        \State Update: $\theta \leftarrow \text{Adam}(\nabla \mathcal{L}_{\text{GRPO}})$
    \EndFor
\EndFor
\State \Return Fine-tuned Artist
\end{algorithmic}
\end{algorithm}

\subsubsection{Artist Training}

Algorithm~\ref{alg:artist_training} presents the complete Artist training with GRPO using four compositional rewards and curriculum learning. This is an extension from the explanation in Section~\ref{sec:artist} of the main text.

\clearpage

\section{Quantitative Results}
\label{app:quant}

This section presents additional quantitative analysis and ablation studies that complement our main paper results. We begin by analyzing the impact of coordinate sorting on Architect-A training stability, followed by standalone Architect performance demonstrating the necessity of task-specific training. We then visualize our main results through a radar chart comparison and present our grid search analysis over reward weight configurations. Next, we provide detailed performance breakdowns across varying numbers of people, revealing how methods scale with scene complexity. Finally, we analyze latency and speed comparisons, demonstrating Ar2Can's superior quality-speed trade-off. These analyses provide deeper insights into the effectiveness of our design choices and training strategies.

\subsection{Data Sorting for Coordinates Regression}
\begin{figure}[h]
  \centering
  \includegraphics[width=0.9\linewidth]{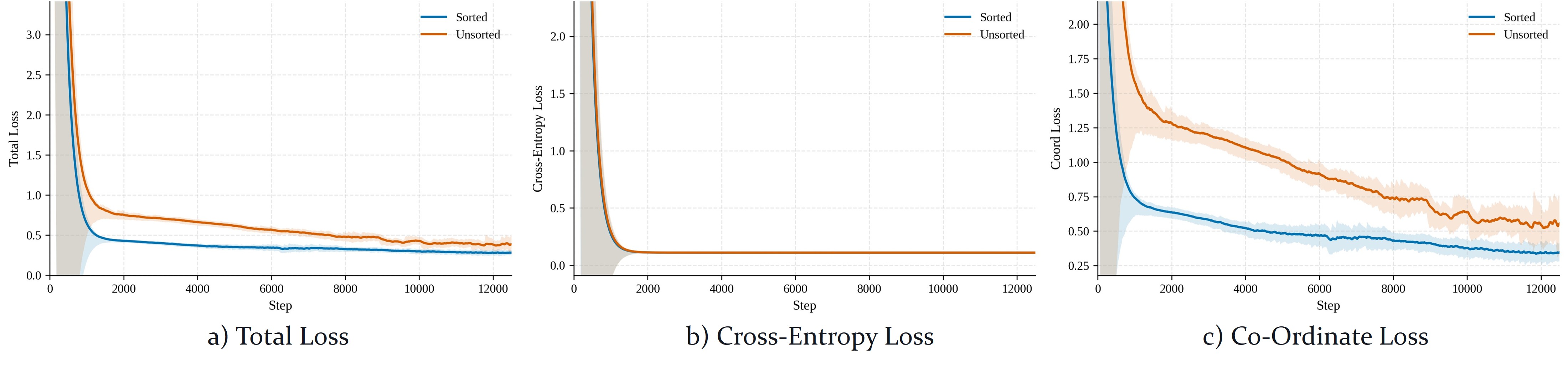}
  \caption{\textbf{Training curves of Architecture-A (LLM-based) with and without data sorting.}
After 2000 steps, training is dominated by the optimization of bounding-box coordinate regression. A clear gap emerges between sorted and unsorted data: sorting leads to more stable learning and faster convergence in the coordinate regression stage under the same number of training iterations.}
  \label{fig:arch_a_curves_comparison_sorted_unsorted}
\end{figure}

In the LLM-based Architect, we observe that the MLP layers are highly sensitive to the permutation of bounding box coordinate embeddings, which negatively affects performance and convergence stability. To preserve simplicity and scalability in lightweight MLP design, we standardize the input by sorting all coordinates in a consistent order (left to right, top to bottom). This simple yet effective bias improves training stability and enables the model to focus more reliably on coordinate regression. The sorted variant exhibits smoother convergence and overall stronger learning behavior compared to the unsorted setting, as shown in Figure~\ref{fig:arch_a_curves_comparison_sorted_unsorted}. Compared to the more mature spatial reasoning in the text-to-image framework of Architecture-B, the LLM-based approach in Architecture-A achieves a decent level of spatial structure, indicating that its coordinate representations can be effectively leveraged for the Artist stage, as shown in Figure~\ref{fig:arch_comparison}.

\subsection{Standalone Architect Performance}

Table~\ref{tab:architect_count} demonstrates the effectiveness of our training procedures for both Architect variants on count accuracy and spatial diversity. We measure \textbf{RMS Spread}, defined as the root mean square of pairwise distances between predicted bounding box centers, which quantifies how well-distributed face locations are across the image (higher values indicate more spatially diverse, natural arrangements). As the prompts differ, a higher spread is ideal. Both Supervised Fine-Tuning (SFT) for Architect-A and Reinforcement Fine-Tuning (RFT) for Architect-B dramatically improve count accuracy over their pretrained baselines. Architect-A achieves 97.7\% (from 15.2\%) and Architect-B reaches 93.2\% (from 59.9\%). Notably, Architect-B produces higher RMS Spread (0.07) than Architect-A (0.05), reflecting its 2D T2I backbone's stronger spatial priors for distributing people naturally across the scene.

\begin{table}[htbp]
    \renewcommand{\arraystretch}{1.5}
    \fontsize{8.0pt}{6.75pt}\selectfont
    \centering
    \begin{tabular}{l|cc}
        \hline
        Model & Count Accuracy$\uparrow$ & RMS Spread$\uparrow$ \\
        \hline
        \multicolumn{3}{c}{\cellcolor{lightyellow}\textbf{MultiHuman-TestBench}} \\
        \hline
        Qwen-2.5-0.5B (Baseline)          & 15.2          & 0.01    \\
        Qwen-2.5-0.5B + SFT (Architect-A) & \cellcolor{green!20}97.7 & 0.05 \\
        \hline
        Flux-Schnell (Baseline)            & 59.9          & 0.04    \\
        Flux-Schnell + RFT (Architect-B)   & \cellcolor{green!10}93.2 & \cellcolor{green!20}0.07 \\
        \hline
    \end{tabular}
    \caption{Count accuracy and spatial diversity (RMS Spread) of Architect variants on MultiHuman-Testbench. RMS Spread measures the root mean square of pairwise distances between predicted bounding box centers; higher values indicate more spatially diverse layouts. Both training procedures significantly improve count accuracy over their baselines, with Architect-B producing more spread-out spatial arrangements owing to its 2D generative backbone.}
    \label{tab:architect_count}
\end{table}

\begin{figure}[h]
    \centering
    \includegraphics[width=0.7\linewidth]{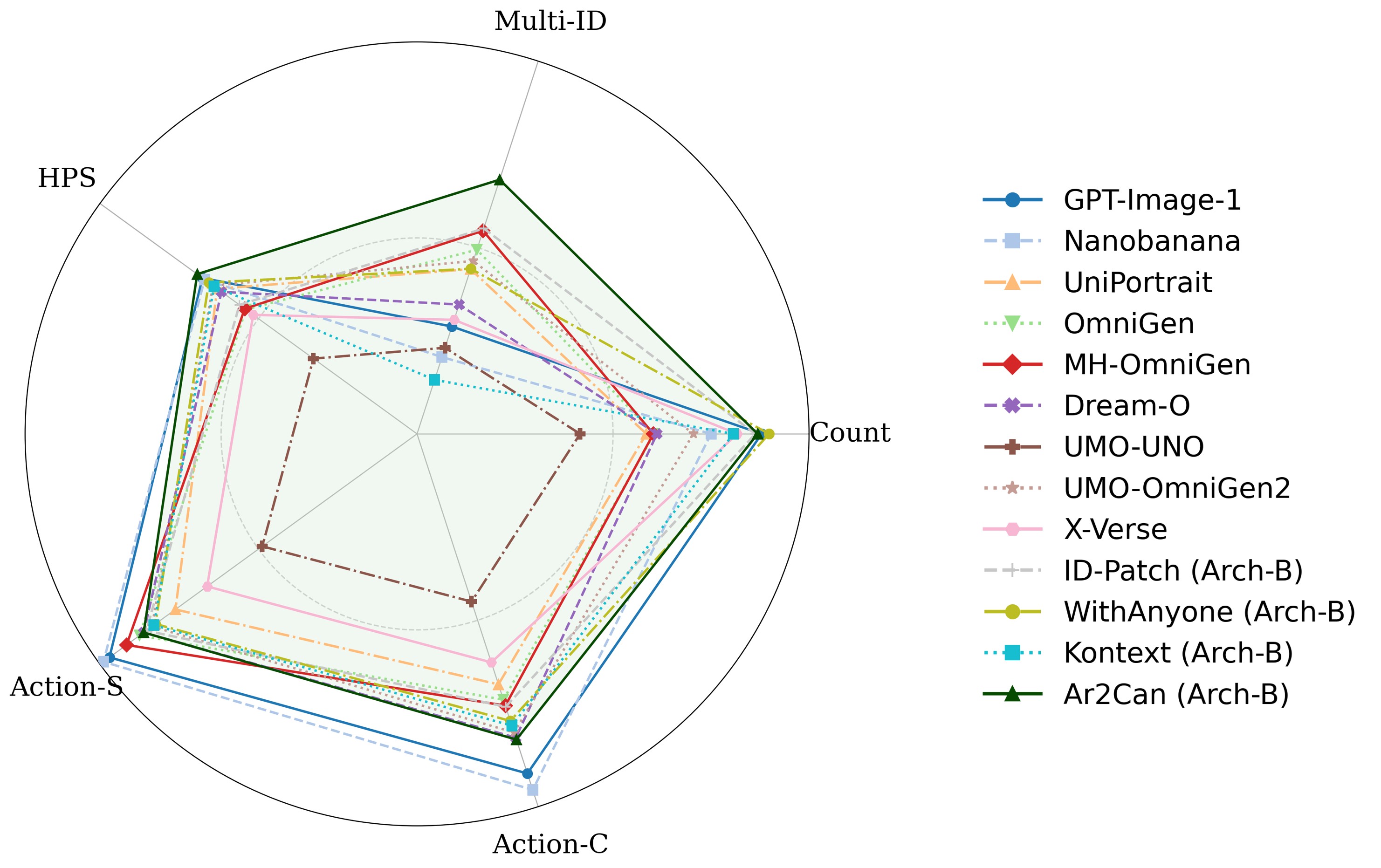}
    \caption{Radar chart visualization of results from Table~\ref{tab:multihuman_recent}. Ar2Can achieves the most balanced performance across all MultiHuman-Testbench metrics, demonstrating superior count accuracy and identity preservation while maintaining competitive prompt alignment and action scores. The SOTA methods exhibit clear trade-offs, excelling in some metrics while failing in others.}
    \label{fig:radar_chart}
\end{figure}

\subsection{Grid Search on Reward Weights}

Table~\ref{tab:artist_ablation} presents our grid search over reward weight configurations. Due to compute constraints, we evaluate at 150 training epochs, for a 7-GPU run. Each reward component offers distinct benefits: \textbf{Face matching} ($\zeta$) provides the strongest identity preservation signal (84.9 Multi-ID when isolated), but sacrifices count accuracy and prompt alignment. The resulting images look poor. We show them in Figure~\ref{fig:hps_impact}. Hence, we need a high quality reward signal such as \textbf{HPSv3} ($\beta$). It is critical for both realism and prompt alignment, with higher weights improving aesthetic quality but potentially overwhelming identity signals. This is also visible in Figure~\ref{fig:hps_impact}. Next, we increase \textbf{Count} reward weight ($\alpha$), which directly improves counting but can conflict with natural scene composition. Finally, \textbf{Frontal Pose} ($\eta$) reduces copy-paste artifacts by encouraging natural face orientations.

As observed in Table~\ref{tab:artist_ablation}, our selected configuration in \colorbox{lightblue}{blue} ($\alpha=0.2, \beta=0.4, \zeta=0.3, \eta=0.1$) achieves the best balance: strong count accuracy (84.5), high identity preservation (65.9), and competitive prompt alignment (29.2). This weighting prioritizes HPSv3 for realism while maintaining sufficient face matching signal for identity preservation, with modest count and pose rewards to prevent failure modes. Extreme configurations (e.g., $\zeta=1.0$) achieve higher single-metric performance but fail to generalize across evaluation dimensions.

\begin{table}[htbp]
    \renewcommand{\arraystretch}{1.5}
    \fontsize{8.0pt}{6.75pt}\selectfont
    \centering
    \begin{tabular}{l|cccc|ccc}
        \hline
        \multirow{2}{*}{\centering Model} & \multicolumn{4}{c|}{Reward Weights} & \multicolumn{3}{c}{Metrics} \\
        \cline{2-8}
          & \makecell{Count\\$\alpha$}
          & \makecell{HPSv3\\$\beta$}
          & \makecell{Face\\$\zeta$}
          & \makecell{Pose\\$\eta$}
          & Count Acc.
          & Multi-ID
          & Prompt Align \\
        \hline
        Baseline & 0 & 0 & 0 & 0 & 80.7 & 14.5 & 29.2 \\
        \hline
        & 0 & 0 & 1 & 0 & 72.7 & \cellcolor{green!20}84.9 & 25.6 \\
        & 0 & 0.5 & 0.5 & 0 & 82.3 & 72.2 & 28.6 \\
        & 0.2 & 0.6 & 0.2 & 0 & \cellcolor{green!20}85.3 & 35.2 & \cellcolor{green!20}29.9 \\
        & 0.2 & 0.3 & 0.3 & 0.2 & 84.3 & 58.2 & 28.4 \\
        \cellcolor{lightblue} Ar2Can & \cellcolor{lightblue} 0.2 & \cellcolor{lightblue} 0.4 & \cellcolor{lightblue} 0.3 & \cellcolor{lightblue} 0.1 & \cellcolor{lightblue} 84.5 & \cellcolor{lightblue} 65.9 & \cellcolor{lightblue} 29.2 \\
        \bottomrule
    \end{tabular}
    \caption{Grid search on Artist reward weights at 150 epochs. Each reward offers distinct benefits: face matching ($\zeta$) maximizes identity preservation, HPSv3 ($\beta$) enhances realism and prompt alignment, count ($\alpha$) improves accuracy, and pose ($\eta$) reduces copy-paste artifacts. Our \colorbox{lightblue}{selected configuration} balances all objectives. Performance differs from main paper due to limited training (fewer epochs/GPUs).}
    \label{tab:artist_ablation}
    \vspace{-5pt}
\end{table}

\subsection{Performance across varying number of people}

Table~\ref{tab:person_count} presents a detailed breakdown of Multi-ID similarity and person count accuracy across different group sizes (2-5 people) on MultiHuman-Testbench. This analysis reveals how methods scale with increasing scene complexity. Most baseline methods exhibit severe performance degradation as the number of people increases, particularly in identity preservation. For instance, MH-OmniGen's Multi-ID drops from 65.3 (2 people) to 38.0 (5 people), while count accuracy falls from 91.2 to 19.7. UniPortrait and OmniGen show similar collapse patterns for larger groups. In contrast, Ar2Can maintains consistently high performance across all person counts, with Multi-ID scores remaining stable (67.7 to 69.5) and count accuracy actually improving with more people (85.1 to 90.9). This demonstrates that our two-stage architecture with explicit spatial grounding effectively prevents the identity merging and counting failures that plague end-to-end methods in complex multi-human scenarios.

\begin{table}[htbp]
    \vspace{1.2 em}
    \renewcommand{\arraystretch}{1.5}
    \fontsize{8.0pt}{6.75pt}\selectfont
    \centering
    \begin{tabular}{l|ccccc|ccccc}
        \hline
        \multirow{2}{*}{Model} & \multicolumn{5}{c|}{Multi-ID Similarity} & \multicolumn{5}{c}{Person Count Accuracy} \\
        & 2 & 3 & 4 & 5 & Avg & 2 & 3 & 4 & 5 & Avg \\
        \hline
        \multicolumn{11}{c}{\cellcolor{lightyellow}\textbf{MultiHuman-Testbench}} \\
        \hline
GPT-Image-1    & 31.8 & 29.5 & 27.8 & 24.9 & 28.8 & 90.7 & 91.8 & 89.5 & 75.3 & 87.9 \\
Nanobanana     & \cellcolor{red!10}21.8 & \cellcolor{red!10}17.5 & \cellcolor{red!10}11.8 & \cellcolor{red!10}10.4 & \cellcolor{red!10}20.6 & \cellcolor{red!10}84.0 & 81.1 & 71.5 & 55.7 & 75.0 \\
\hline
Fastcomposer   & \cellcolor{red!20}15.3 & \cellcolor{red!20}7.4 & \cellcolor{red!20}7.2 & \cellcolor{red!20}5.9 & \cellcolor{red!20}12.2 & \cellcolor{red!20}62.9 & \cellcolor{red!20}11.2 & \cellcolor{red!20}3.2 & \cellcolor{red!20}1.1 & \cellcolor{red!20}31.2 \\
UniPortrait    & 56.5 & 46.4 & 33.8 & 28.6 & 44.2 & 90.6 & \cellcolor{red!10}76.3 & 23.7 & \cellcolor{red!10}14.1 & \cellcolor{red!10}58.5 \\
OmniGen        & 60.8 & 52.3 & 42.2 & 35.2 & 49.4 & 88.8 & 88.0 & 23.2 & 21.6 & 60.5 \\
MH-OmniGen     & \cellcolor{green!10}65.3 & \cellcolor{green!10}60.4 & \cellcolor{green!10}45.1 & 38.0 & \cellcolor{green!10}54.5 & 91.2 & 87.5 & \cellcolor{red!10}22.4 & 19.7 & 60.3 \\
Dream-O        & 48.7 & 30.9 & 20.0 & 15.9 & 34.7 & \cellcolor{green!10}92.3 & 86.1 & 26.9 & 15.2 & 61.2 \\
UMO-OmniGen2   & 56.6 & 49.2 & 36.8 & 30.6 & 46.4 & 92.0 & \cellcolor{green!10}93.6 & 41.3 & 40.0 & 70.5 \\
X-Verse        & 39.6 & 33.9 & 24.6 & 20.7 & 30.6 & \cellcolor{green!20}96.9 & \cellcolor{green!20}95.1 & \cellcolor{green!20}90.1 & 40.0 & 81.7 \\
WithAnyone     & 45.7 & 44.4 & 44.0 & \cellcolor{green!10}42.7 & 44.3 & 90.4 & 91.5 & 89.6 & \cellcolor{green!10}88.3 & \cellcolor{green!10}89.8 \\
\hline
\textbf{Ar2Can (Ours)} & \cellcolor{green!20}67.7 & \cellcolor{green!20}71.8 & \cellcolor{green!20}71.7 & \cellcolor{green!20}69.5 & \cellcolor{green!20}67.6 & 88.1 & 86.9 & \cellcolor{green!10}89.9 & \cellcolor{green!20}90.9 & \cellcolor{green!20}90.2 \\
        \hline
    \end{tabular}
    \caption{Performance breakdown across different group sizes (2-5 people) on MultiHuman-Testbench. Color coding: \colorbox{green!20}{highest} and \colorbox{red!20}{lowest}. Ar2Can (Architect-A) maintains consistently high Multi-ID similarity and count accuracy across all person counts, while baseline methods exhibit severe degradation with increasing scene complexity.}
    \label{tab:person_count}
\end{table}
\subsection{Latency Analysis and Speed Comparisons}

Table~\ref{tab:latency} provides a detailed breakdown of inference times for Ar2Can and state-of-the-art methods on A100 GPU at $1024 \times 1024$ resolution with 3 identities. We separately report the Architect inference time (layout generation), Artist inference time (image rendering), and total latency for our two-stage approach.

As observed, Ar2Can achieves the best quality-speed trade-off among all methods. Architect-A (Qwen-based) requires only 0.5 seconds for layout generation, while Architect-B (Flux-Schnell based) takes 1.4 seconds. The Artist with token sharing reduces inference time by approximately 2$\times$ compared to the full canvas baseline (15s vs 28s), while maintaining the highest unified scores (72.4 and 72.2). Architect-A offers the fastest overall latency (15.5s) with the highest unified score (72.4), while Architect-B provides slightly lower but still competitive performance (72.2) with a total latency of 16.4s.

Compared to other methods, Ar2Can with token sharing (Arch-A) achieves the second fastest total latency (15.5s) among all methods while delivering significantly higher quality (72.4 unified score). Methods like UMO-OmniGen2 and X-Verse require substantially longer inference times (74s and 87s respectively) while achieving lower unified scores. This demonstrates that our modular architecture with efficient token sharing provides superior performance without sacrificing speed.

\begin{table}[htbp]
    \renewcommand{\arraystretch}{1.5}
    \setlength{\tabcolsep}{6pt} 
    \fontsize{8.5pt}{7pt}\selectfont
    \centering
    \begin{tabular}{l|c|cccc}
        \hline
        Model & Diffusion Steps & Architect (sec) & Artist (sec) & Total (sec) $\downarrow$ & Unified Score $\uparrow$ \\
        \hline
        MH-OmniGen & 50 & — & 59 & 59 & 61.6 \\
        UMO-OmniGen2 & 50 & — & 74 & \cellcolor{red!10}74 & 60.4 \\
        Dream-O & 25 & — & 57 & 57 & \cellcolor{red!10}59.7 \\
        X-Verse & 50 & — & 87 & \cellcolor{red!20}87 & \cellcolor{red!20}52.7 \\
        \hline
        WithAnyone (Arch-B) & 25 & 1.4 & 14 & \cellcolor{green!20}15.4 & 62.6 \\
        Ar2Can-Full (Arch-B) & 28 & 1.4 & 28 & 29.4 & 71.5 \\
        Ar2Can-Shared (Arch-A) & 28 & 0.5 & 15 & \cellcolor{green!10}15.5 & \cellcolor{green!20}72.4 \\
        Ar2Can-Shared (Arch-B) & 28 & 1.4 & 15 & 16.4 & \cellcolor{green!10}72.2 \\
        \bottomrule
    \end{tabular}
    \caption{Latency breakdown on A100 GPU ($1024 \times 1024$, 3 identities). Color coding: \colorbox{green!20}{best} and \colorbox{red!20}{lowest}. Arrows indicate optimization direction: $\downarrow$ lower is better, $\uparrow$ higher is better. Ar2Can's Architect-A achieves the fastest total latency (15.5s) with the highest unified score (72.4), demonstrating superior quality-speed trade-off. Token sharing accelerates the Artist by approximately 2$\times$ (15s vs 28s full canvas).}
    \label{tab:latency}
\end{table}

\subsection{Unified Architecture Variant (Architect-C)}
\label{app:unified_arch}

To validate our choice of two lightweight specialist Architects, we introduce \textbf{Architect-C}, which finetunes BAGEL~\cite{deng2025emerging}, which is a unified multimodal large language model. For this finetuning, we use the same GRPO procedure as Architect-B (Section~\ref{sec:architect_b}). Table~\ref{tab:unified_arch} compares all three variants across standalone layout quality and final Artist performance. Architect-C achieves competitive results (ID: 68.0, HPS: 30.9, RMS Spread: 0.09) comparable to Architect-A and Architect-B, confirming that a unified model can capture both language understanding and spatial reasoning under our RL training. It obtains the best RMS spread and HPS which signifies that this has the best spatial layout compared to other methods. However, it incurs a 27s layout generation time. This is roughly 19$\times$ slower than Architect-A (0.5s) and Architect-B (1.4s), making it impractical for deployment despite marginally higher layout diversity. This confirms that our two lightweight specialists provide a superior efficiency v/s quality tradeoff.

\begin{table}[h]
    \renewcommand{\arraystretch}{1.5}
    \fontsize{8.0pt}{6.75pt}\selectfont
    \centering
    \begin{tabular}{l|ccc|ccc}
        \hline
        \multirow{2}{*}{Architect} & \multicolumn{3}{c|}{Architect Performance} & \multicolumn{3}{c}{Our Artist (Final)} \\
        & Acc.$\uparrow$ & RMS Spread$\uparrow$ & Time$\downarrow$ & Acc.$\uparrow$ & ID$\uparrow$ & HPS$\uparrow$ \\
        \hline
        Architect-A           & 97.7 & 0.05 & \cellcolor{green!20}0.5s  & \cellcolor{green!20}90.2 & 67.6          & 30.2          \\
        Architect-B (Schnell) & 93.2 & 0.07 & 1.4s                      & 86.9          & \cellcolor{green!20}68.2 & 30.8 \\
        Architect-C (BAGEL)   & \cellcolor{green!20}96.1 & \cellcolor{green!20}0.09 & \cellcolor{red!20}27s & 88.2 & 68.0 & \cellcolor{green!20}30.9 \\
        \hline
    \end{tabular}
    \caption{Comparison of Architect variants on standalone layout quality and final Artist performance. \textbf{Spread}: RMS spread of bounding box centers (higher = more spatially diverse layouts). Color coding: \colorbox{green!20}{best} and \colorbox{red!20}{worst} per column.}
    \label{tab:unified_arch}
\end{table}

\subsection{Off-the-Shelf Layout Generation as Architect}
\label{app:offtheshelf_arch}

As an alternative to task-specific Architect training, one could use off-the-shelf layout generation methods such as LayoutGPT~\cite{feng2023layoutgpt} and RPG-DiffusionMaster~\cite{yang2024mastering} as drop-in Architects, feeding their predicted layouts directly to our Artist. We use the method proposed by these works to generate boxes, using both GPT-4o and Llama prompted to generate boxes. The results in Table~\ref{tab:offtheshelf_arch} show that while RPG/LayoutGPT with GPT-4o achieves high count accuracy (100.0), both methods produce very similar layouts for every prompt (RMS-Spread: 0.01-0.02). This directly degrades final Artist performance (ID: 57.2 for GPT-4o vs.\ 67.6 for Architect-A), demonstrating that layout \textit{diversity} is as critical as count accuracy, and that task-specific Architect training is necessary.

\begin{table}[h]
    \renewcommand{\arraystretch}{1.5}
    \fontsize{8.0pt}{6.75pt}\selectfont
    \centering
    \begin{tabular}{l|l|ccc|ccc}
        \hline
        \multirow{2}{*}{Architect} & \multirow{2}{*}{Backbone} & \multicolumn{3}{c|}{Architect Performance} & \multicolumn{3}{c}{Our Artist (Final)} \\
        & & Acc.$\uparrow$ & Spread$\uparrow$ & Time$\downarrow$ & Acc.$\uparrow$ & ID$\uparrow$ & HPS$\uparrow$ \\
        \hline
        RPG/LayoutGPT & GPT-4o  & \cellcolor{green!20}100.0 & \cellcolor{red!20}0.01 & 14s & 78.6 & 57.2 & 27.5 \\
        RPG/LayoutGPT & Llama   & \cellcolor{red!20}51.0   & 0.02                  & \cellcolor{green!20}3s  & \cellcolor{red!20}45.3 & \cellcolor{red!20}31.3 & \cellcolor{red!20}27.0 \\
        \hline
        Architect-A & Qwen-2.5-0.5B   & 97.7 & 0.05                  & 0.5s & \cellcolor{green!20}90.2 & 67.6                  & 30.2                  \\
        Architect-B & Flux-Schnell    & 93.2 & \cellcolor{green!20}0.07 & 1.4s & 86.9          & \cellcolor{green!20}68.2 & \cellcolor{green!20}30.8 \\
        \hline
    \end{tabular}
    \caption{Off-the-shelf layout generators vs.\ our trained Architects as drop-in replacements feeding the same Artist. Despite high count accuracy, RPG/LayoutGPT produces low-diversity layouts (Spread) that substantially degrade final generation quality. Color coding: \colorbox{green!20}{best} and \colorbox{red!20}{worst} per column.}
    \label{tab:offtheshelf_arch}
\end{table}

\section{Qualitative Results}
\label{app:qual}

This section presents comprehensive qualitative analysis and visual comparisons that illustrate the effectiveness of our proposed components. We begin with extensive qualitative comparisons on the Multi-ID Test benchmark, demonstrating Ar2Can's ability to jointly optimize identity preservation and prompt alignment. We then showcase results from our optional pose-controlled Artist variant and visualize the effectiveness of our token sharing strategy for handling overlapping regions. Additional comparisons demonstrate the advantages of Hungarian centroid matching over naive spatial matching, the complementary strengths of our two Architect variants, the validity of our frontal pose scoring, and the critical role of HPSv3 in enhancing aesthetic quality.

\begin{figure}[h]
    \centering
    \includegraphics[width=\linewidth]{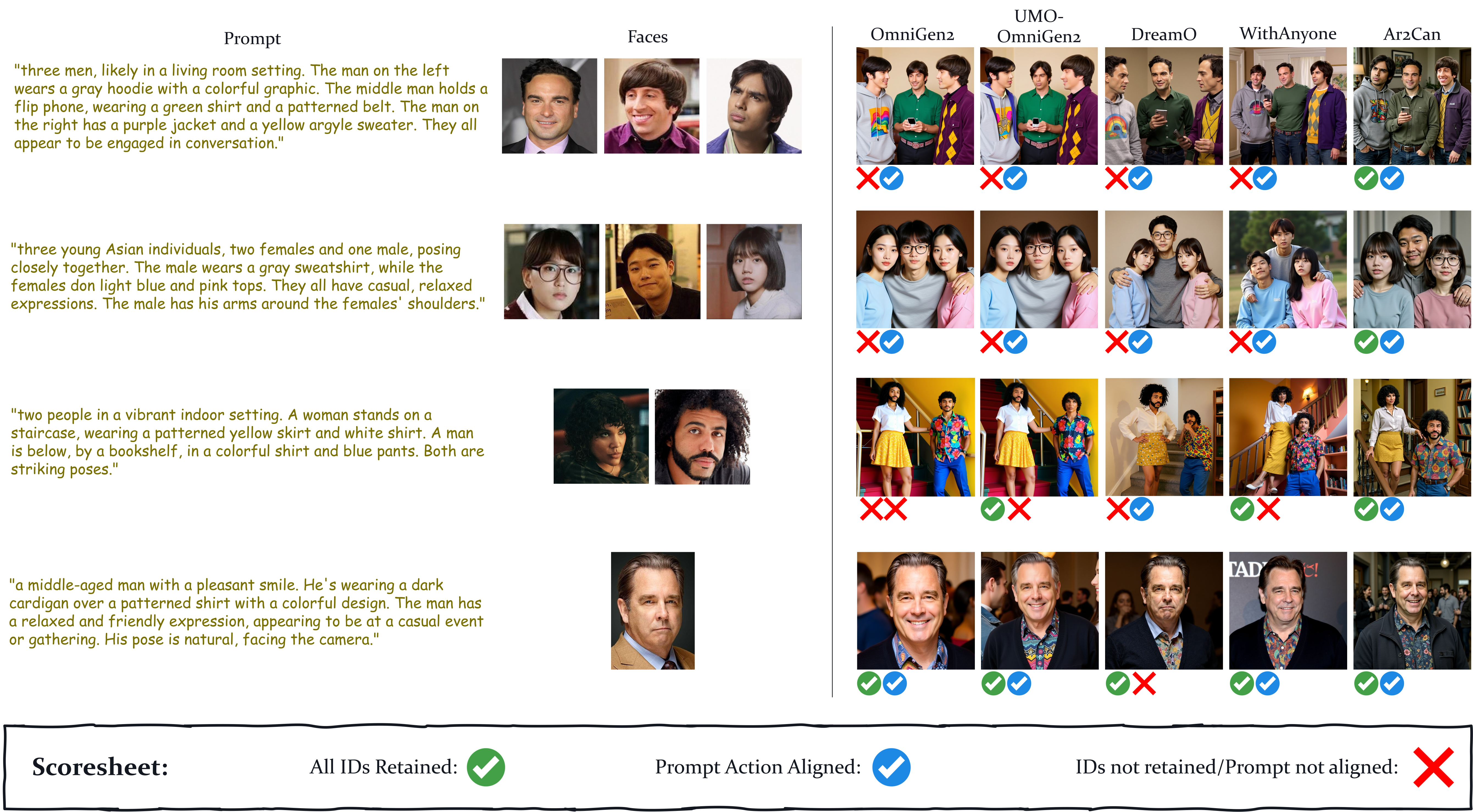}
    \caption{Qualitative comparison with state-of-the-art methods on Multi-ID Test. Each row shows reference identities and generations from multiple methods given highly descriptive prompts specifying clothing, expressions, item placement, and spatial arrangements. Ar2Can consistently preserves all input identities while accurately following detailed prompt specifications. Baseline methods exhibit a trade-off: either achieving identity preservation at the cost of prompt alignment, or vice versa. This limitation becomes more severe as the number of people increases.}
    \label{fig:withanyone_comparison}
\end{figure}

\subsection{Qualitative Comparison on Multi-ID Test}

Figure~\ref{fig:withanyone_comparison} presents a comprehensive qualitative comparison between Ar2Can and state-of-the-art methods on the Multi-ID Test benchmark. Each row shows a different multi-person scenario with highly descriptive prompts that specify fine-grained details including clothing styles, facial expressions, item placement, and spatial arrangements.

As observed across all examples, Ar2Can consistently achieves both objectives simultaneously: (1) faithful identity preservation for every input face, and (2) accurate alignment with detailed prompt specifications. In contrast, existing methods exhibit a clear trade-off between these objectives. Methods optimized for identity preservation often fail to execute prompt-specified details correctly. They miss clothing attributes, incorrect expressions, or wrong item placements. Conversely in cases where methods achieve better prompt alignment, frequently suffer from identity hallucination or blending.

This failure to jointly optimize both objectives becomes more pronounced as the number of people increases. For scenes with 3 people, baseline methods systematically fail at either identity consistency or prompt adherence. Ar2Can has a two-stage architecture, and the artist's GRPO-based training method contains explicit spatial grounding and Hungarian-based face matching. This enables it to maintain both high identity fidelity and precise prompt alignment across varying person counts.

\begin{figure}[h]
    \centering
    \includegraphics[width=0.7\linewidth]{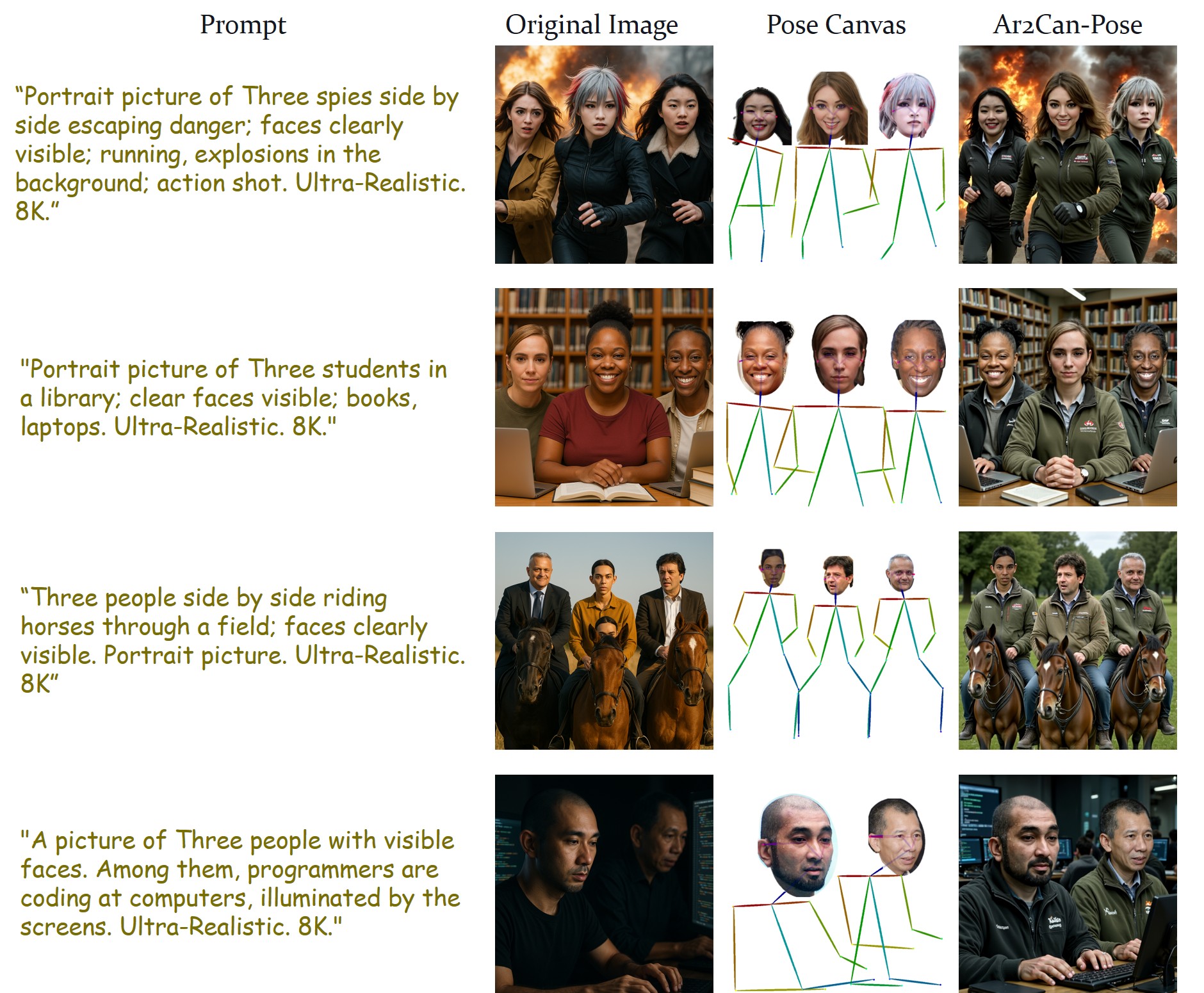}
    \caption{Results from our pose-controlled Artist variant trained with pose rewards. The model successfully preserves all input identities while accurately aligning with specified input poses, generating photorealistic multi-person scenes with controlled actions and body configurations.}
    \label{fig:posemode_results}
\end{figure}

\subsection{Pose-Controlled Artist Results}

Figure~\ref{fig:posemode_results} demonstrates the capabilities of our pose-controlled Artist variant, trained with the pose reward described in Section~\ref{app:method}. This variant conditions on both face bounding boxes and human pose skeletons overlaid on the canvas, enabling explicit control over body poses and actions in the generated images.

As observed in the results, the pose-controlled Artist successfully achieves multiple objectives: (1) faithful identity preservation for all input faces across different scenes, (2) accurate alignment with the specified input poses, and (3) photorealistic rendering quality with natural lighting and coherent scene composition. The model demonstrates the ability to generate diverse multi-person scenarios while maintaining precise pose control. It can generate images ranging from complex group poses to dynamic action sequences, all while preserving the distinctive facial features of each reference identity.

However, we observe an increase in copy-paste artifacts in this variant compared to our standard Artist. The additional constraint of matching specific body poses appears to occasionally lead to more rigid spatial composition, where faces and bodies are sometimes rendered with less natural integration into the scene context. This represents a trade-off between pose controllability and rendering flexibility. We note that these results represent an initial exploration of pose-controlled multi-human generation. The pose-controlled Artist demonstrates the feasibility of fine-grained action control while maintaining identity preservation, but addressing the copy-paste artifacts requires further investigation. We are actively working on improving this variant. This direction represents promising future work for extending Ar2Can's capabilities to fine-grained controllable generation.

\begin{figure}[h]
    \centering
    \includegraphics[width=0.7\linewidth]{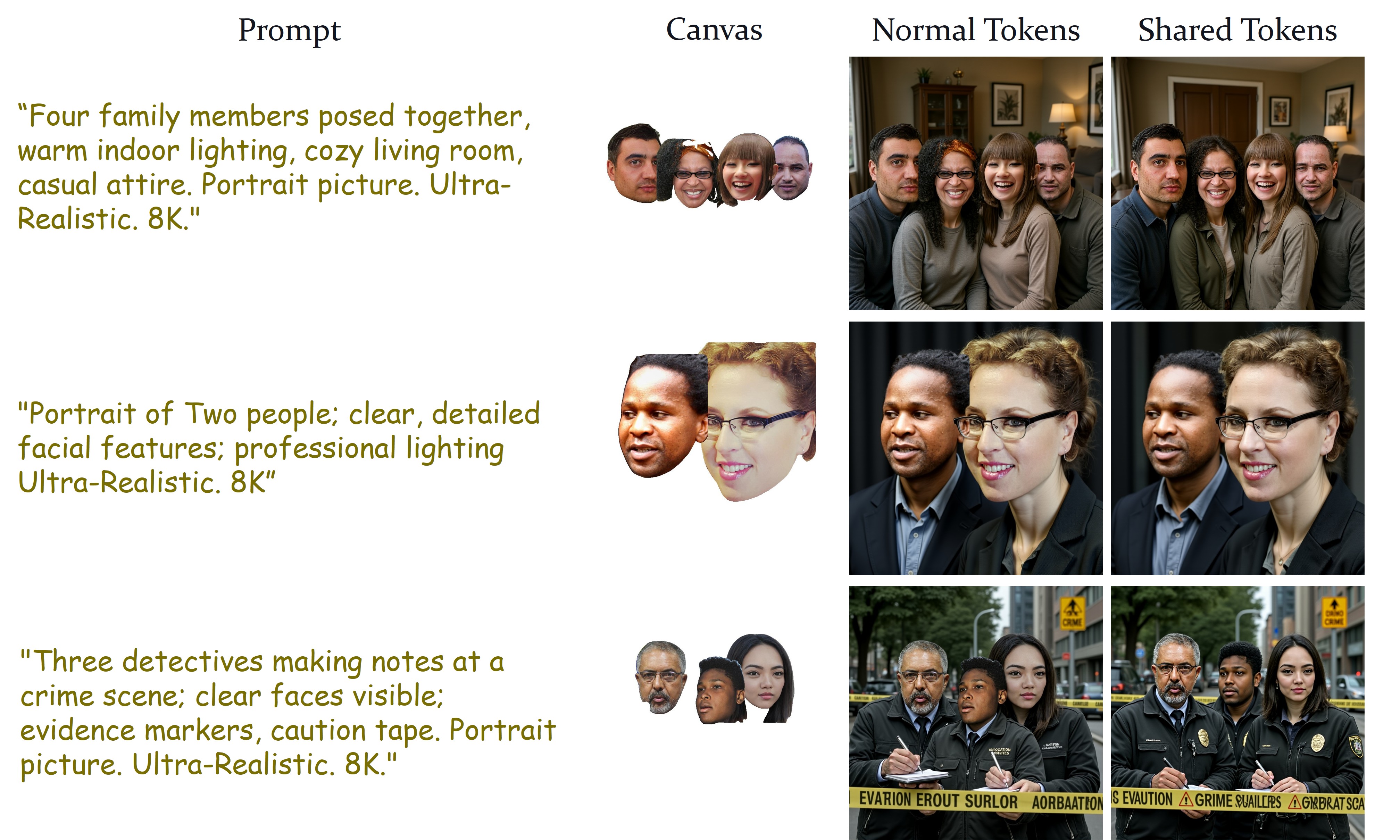}
    \caption{Visual comparison demonstrating the effectiveness of token sharing. \textbf{Left}: Canvas input with overlapping face regions. \textbf{Middle}: Generation without shared tokens, resulting in unnatural spatial conflicts. \textbf{Right}: Generation with shared tokens, enabling natural depth ordering and realistic occlusions. The shared token approach produces perceptually superior results while maintaining controllability when explicit depth ordering is required.}
    \label{fig:token_sharing_results}
\end{figure}

\subsection{Effectiveness of Token Sharing}

In Figure~\ref{fig:token_sharing_results}, we demonstrate the effectiveness of our proposed token sharing strategy for handling overlapping spatial regions. The figure shows four columns: the prompt, canvas input, generation without shared tokens, and generation with shared tokens.

As observed in the results, the shared token approach enables the model to naturally determine depth ordering between people. When multiple face regions overlap, the model learns to resolve spatial conflicts through intelligent compositional strategies. It either arranges people with appropriate depth layering or by spatially reorganizing them to avoid unnatural overlap. Hence, the results with shared tokens are perceptually better, exhibiting more natural poses, realistic occlusions, and coherent spatial arrangements. Importantly, this approach is both automatic and controllable. When no explicit ordering is desired, the shared positional encodings allow the model to determine the most natural composition. However, \textbf{when specific depth control is needed} (e.g., to place a particular person in the foreground), we can selectively disable shared encodings for the region corresponding to the person who should appear behind. This provides users with fine-grained control over depth ordering while maintaining the perceptual benefits of natural scene composition.

\begin{figure}[h]
    \centering
    \includegraphics[width=0.4\linewidth]{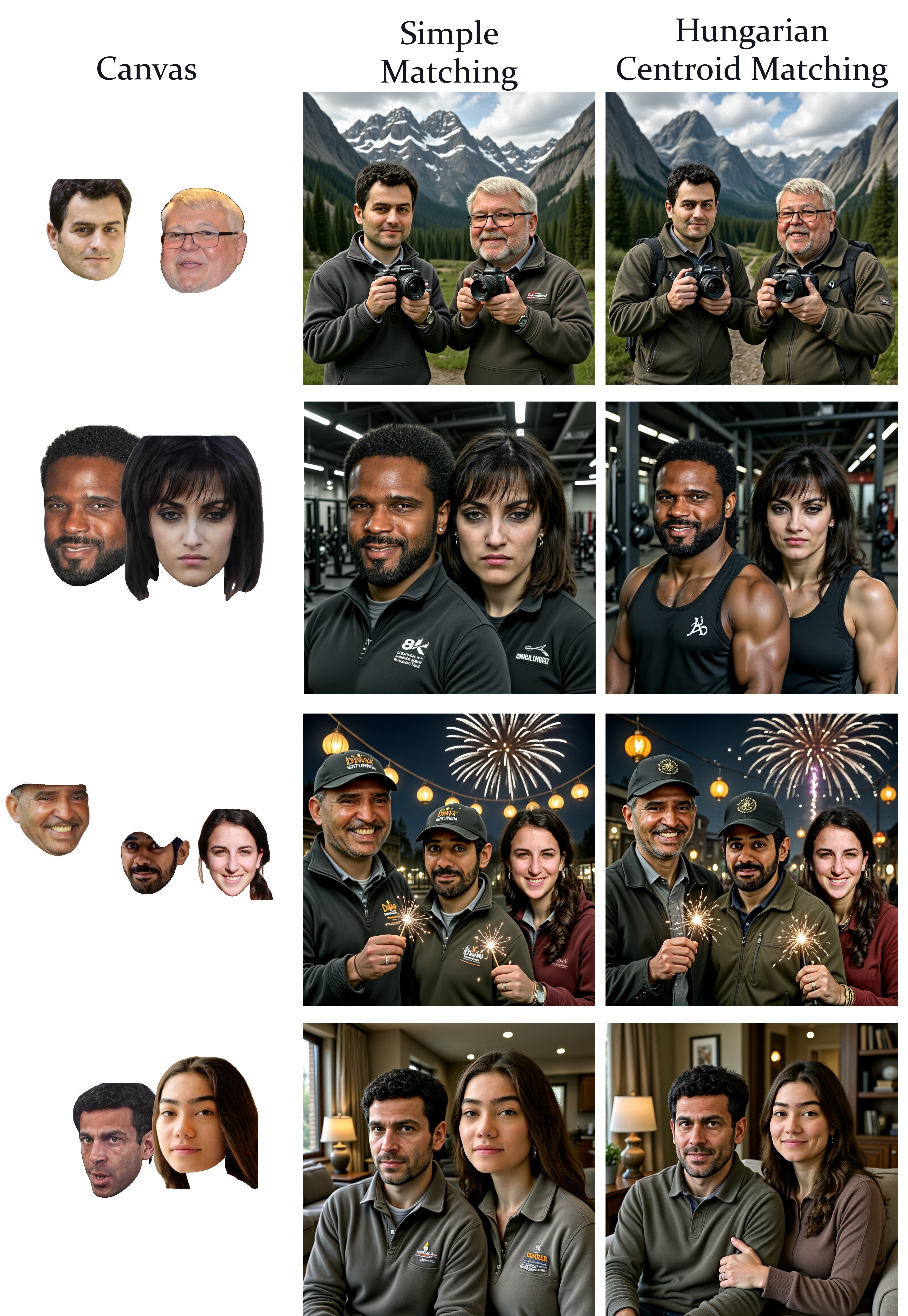}
    \caption{Visual comparison between naive location-based matching and Hungarian Centroid Matching (HCM). \textbf{Simple Matching} enforces exact spatial localization, leading to copy-paste artifacts and unnatural face sizes that degrade perceptual quality. \textbf{Hungarian Centroid Matching} relaxes spatial constraints by matching on centroid proximity, enabling the model to flexibly adjust face sizes and positions based on prompt semantics and aesthetic considerations. This produces photorealistic results with natural proportions while maintaining strong identity preservation.}
    \label{fig:matching_comparison}
\end{figure}

\subsection{Naive Face Matching vs. Hungarian Centroid Face Matching}

Figure~\ref{fig:matching_comparison} provides visual evidence for the effectiveness of our Hungarian Centroid Matching (HCM) approach compared to naive location-based matching. As shown in Table 3 of the main paper, simple matching improves Multi-ID scores (55.2) but significantly degrades count accuracy (80.7→75.6) and image quality (HPS: 29.2→27.6).

The visual results clearly illustrate the underlying problem: simple matching, which extracts faces at exact predicted bounding box locations, produces copy-paste artifacts and faces with unnatural sizes. The rigid constraint of exact spatial localization forces the model to paste reference faces directly at specified coordinates, often resulting in faces that are too large, too small, or incorrectly scaled relative to the body and scene context. These artifacts create perceptually jarring images that lack photorealism despite achieving reasonable identity similarity.

In contrast, Hungarian Centroid Matching relaxes these rigid spatial constraints by matching faces based on centroid proximity rather than exact bounding box overlap. This flexibility allows the model to adjust face sizes and precise locations based on prompt requirements and aesthetic considerations. The model can now render faces with natural scaling, appropriate depth cues, and realistic proportions relative to bodies and the overall scene composition. As demonstrated in the ablation study (Table 3), HCM recovers image quality (HPS: 30.9) while further improving Multi-ID scores (60.3). Hence, it achieves the best balance between spatial accuracy, identity preservation, and photorealism.

\begin{figure}[h]
    \centering
    \includegraphics[width=0.7\linewidth]{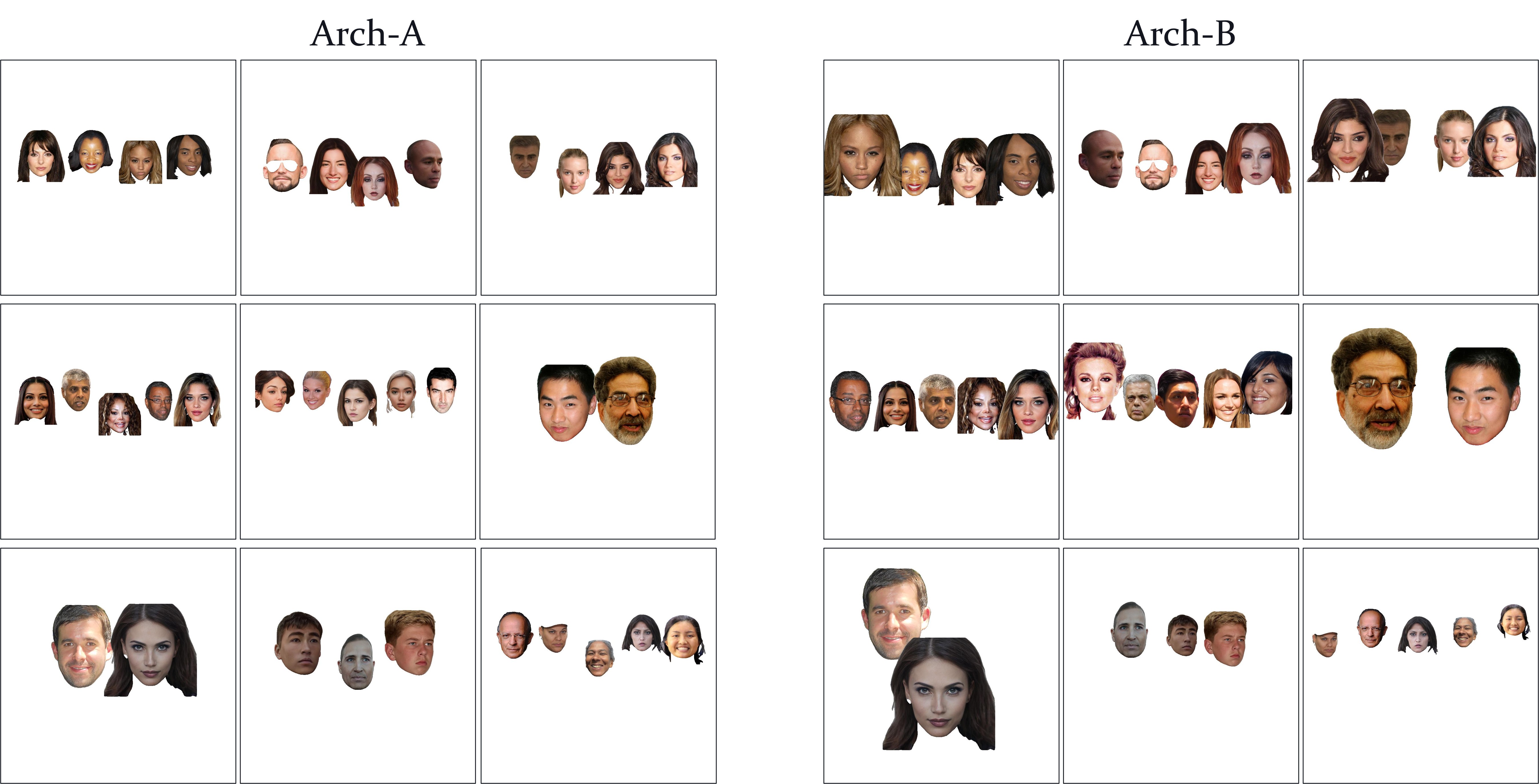}
    \caption{Visual comparison of canvases generated by Architect-A (left) and Architect-B (right) given identical prompts. Architect-B produces more spatially diverse layouts with spread out face arrangements and varied distributions, leveraging its 2D T2I backbone to capture complex spatial relationships. Architect-A generates more compact, structured layouts with superior count accuracy and faster inference times. Each variant offers complementary strengths for different deployment scenarios.}
    \label{fig:arch_comparison}
\end{figure}

\subsection{Architect Variant Comparison}

Figure~\ref{fig:arch_comparison} presents a visual comparison of canvases generated by our two Architect variants given identical text prompts. The figure illustrates the fundamental differences in spatial layout generation between Architect-A (LLM-based, left) and Architect-B (T2I-based, right).

As observed in the results, Architect-B produces more spread out face arrangements with greater spatial diversity in the distribution of people across the canvas. The 2D nature of the T2I backbone enables it to naturally capture complex spatial relationships and varied compositional layouts, often placing people at different depths and with more dynamic spatial configurations. This results in naturally varied scene compositions.

In contrast, Architect-A generates layouts with more compact and structured face arrangements. While these layouts may appear less spatially diverse, Architect-A demonstrates superior count accuracy due to its strong language understanding capabilities, which enable it to reliably parse numerical references from text prompts. Additionally, Architect-A achieves significantly faster inference times (1.4s vs 0.5s as shown in Table~\ref{tab:latency}), making it more suitable for latency-sensitive applications.

Each Architect variant offers distinct advantages: Architect-B excels at generating spatially rich and varied layouts with natural pose information, while Architect-A provides more reliable count accuracy with faster generation speeds. This modular design allows users to select the appropriate Architect based on their specific requirements.

\subsection{Visualizing Frontal Pose Scores}

\begin{figure}[h]
    \centering
    \includegraphics[width=0.7\linewidth]{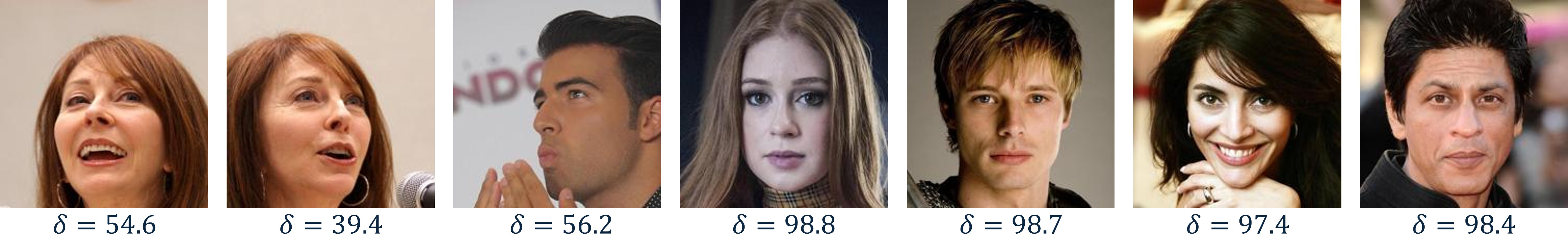}
    \caption{Frontal pose score validation. The keypoint-based scoring method assigns high scores ($>0.9$) to forward-facing faces and low scores to rotated or tilted faces, proving as an effective reward for encouraging natural camera-facing poses during training.}
    \label{fig:frontal_validation}
\end{figure}

Figure~\ref{fig:frontal_validation} illustrates our frontal pose scoring on sample faces from the test set. The frontal score effectively distinguishes frontal faces (score $>0.9$) from non-frontal faces (significantly lower scores). Hence, we use it as a reward signal to reduce copy-paste artifacts.

\subsection{HPSv3 Reward Impact}

\noindent\textbf{Impact on Architect-B.} In Architect-B, removing HPSv3 from the GRPO reward causes the model to reward-hack the count objective by predicting repetitive, clustered layouts regardless of the input prompt. Similar to the analysis shown in Table~\ref{app:offtheshelf_arch}, RMS Spread drops from 0.07 to 0.02, producing very similar spatial arrangements in every image. HPSv3 prevents this by enforcing prompt-aware, spatially diverse layouts.

\begin{figure}[h]
    \centering
    \includegraphics[width=0.8\linewidth]{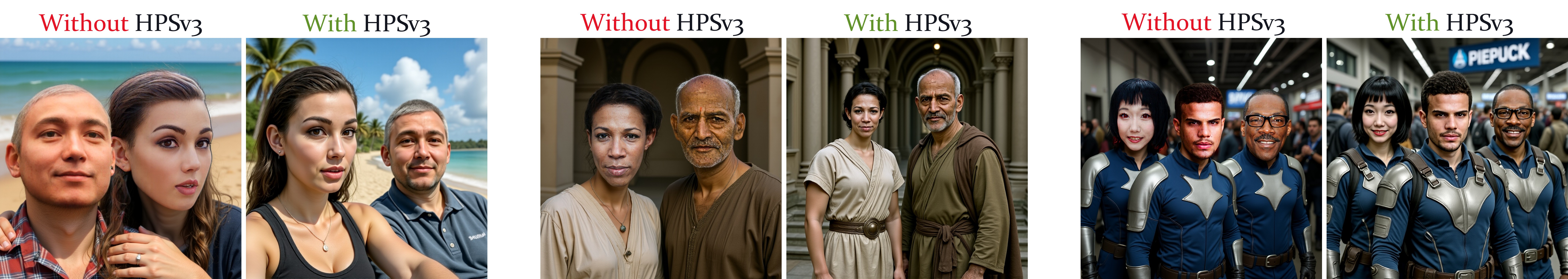}
    \caption{Impact of HPSv3 reward on generation quality. Without HPSv3, images exhibit flat lighting, unnatural colors, and poor compositional balance despite preserving identities.}
    \label{fig:hps_impact}
\end{figure}

\noindent\textbf{Impact on Artist.} Figure~\ref{fig:hps_impact} demonstrates the dual benefit of the HPSv3 reward in Artist training. Beyond improving prompt alignment, HPSv3 significantly enhances overall aesthetic quality, producing more photorealistic lighting, natural skin tones, coherent scene composition, and professional-grade rendering. Visual comparison shows that without HPSv3, generated images suffer from flat lighting, unnatural colors, and poor compositional balance, even when identity preservation is maintained. This validates HPSv3 as a critical component for achieving both semantic accuracy and visual realism in multi-human generation.

\subsection{Generalization to Multi-Object and Multi-Human+Object Scenes}
\label{app:generalization}

\begin{figure}[h]
    \centering
    \includegraphics[width=0.5\linewidth]{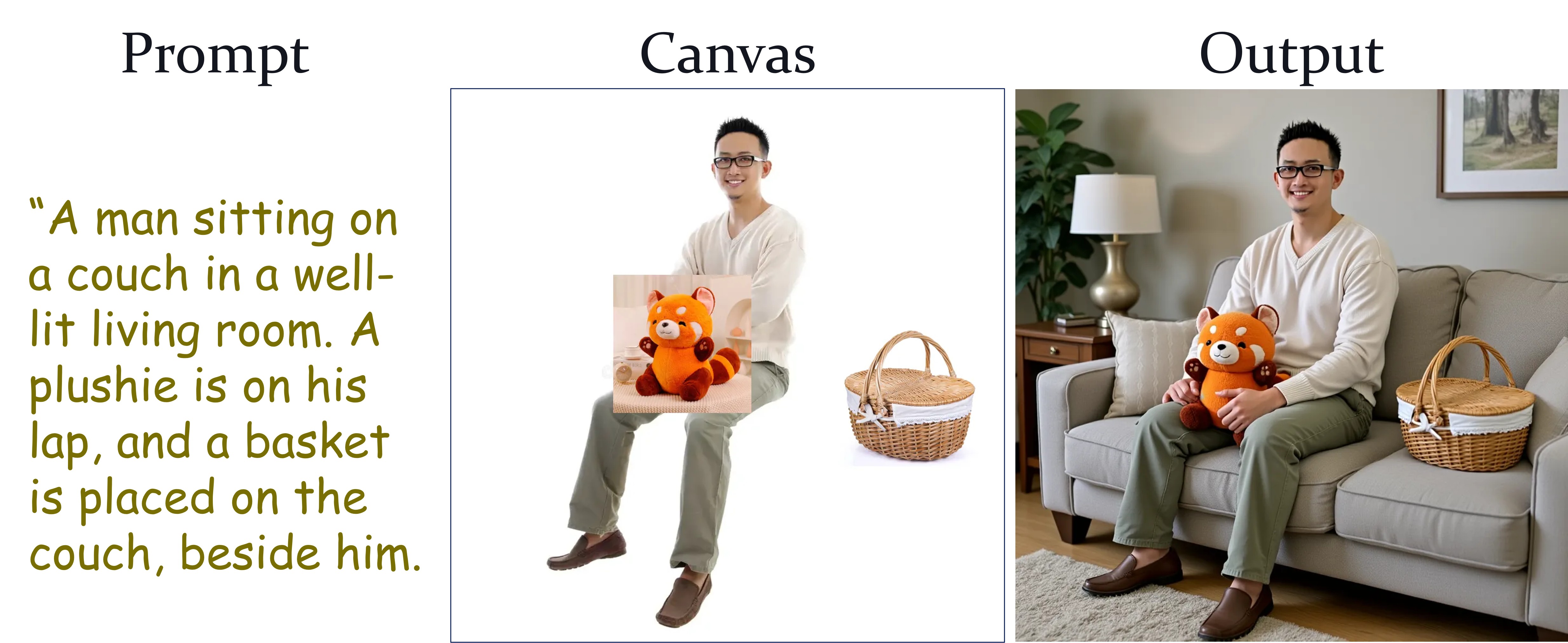}
    \caption{Occlusion handling in multi-object scenes. Given an input prompt, canvas, and output image, our token sharing scheme with shared RoPE encodings naturally resolves spatial conflicts between overlapping objects, producing realistic depth ordering and occlusions without explicit supervision.}
    \label{fig:multiobject_occlusion}
\end{figure}

\begin{figure}[h]
    \centering
    \includegraphics[width=0.6\linewidth]{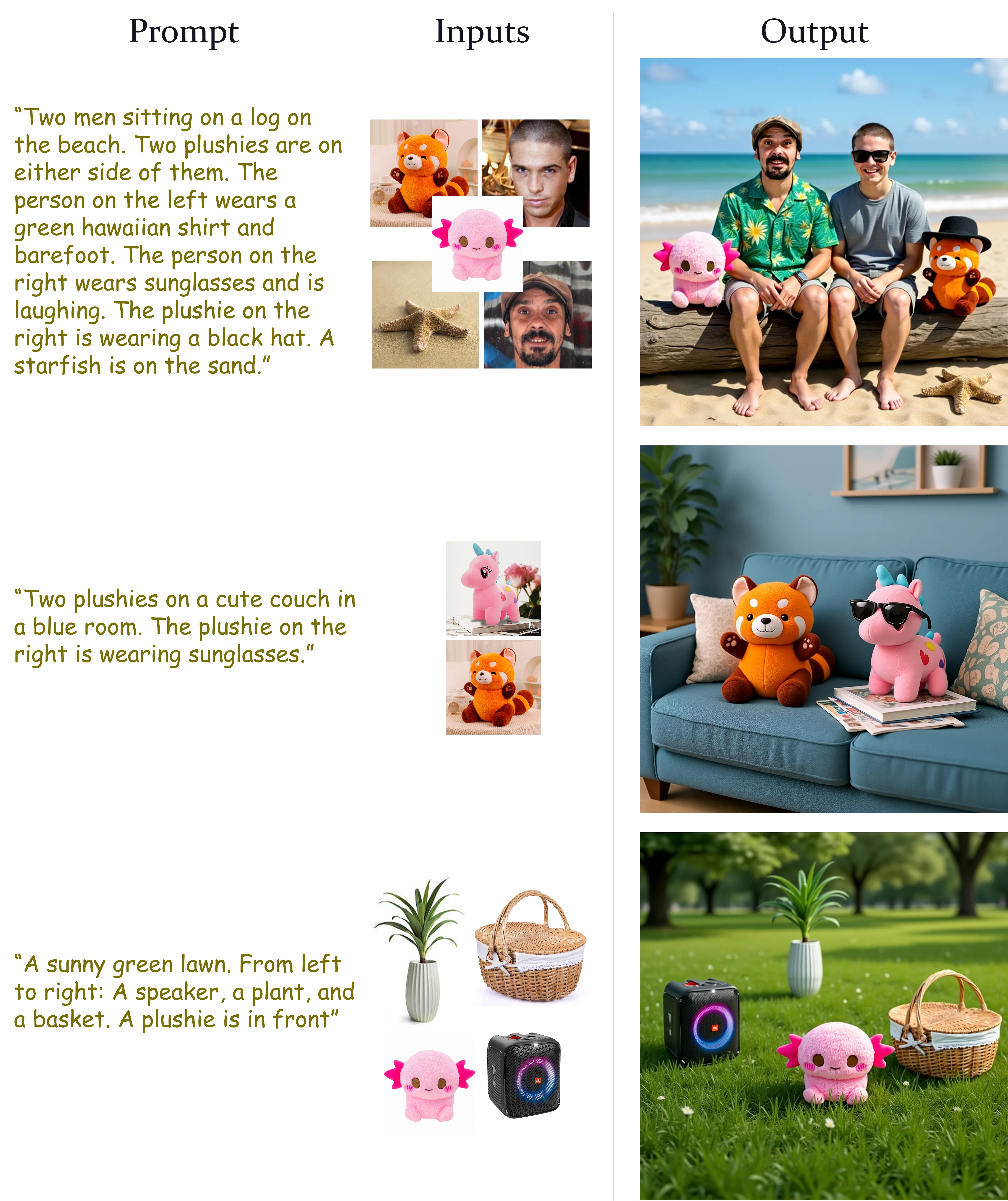}
    \caption{Multi-human+object generation. Given an input prompt and reference images for both people and objects, we construct a canvas by pasting segmented references at Architect-predicted locations. The Artist synthesizes photorealistic outputs preserving all input identities while naturally composing humans and objects within the scene.}
    \label{fig:multiobject_general}
\end{figure}

A notable property of our Artist is its ability to generalize beyond multi-human generation to multi-object and mixed multi-human+object scenes, \textbf{without any retraining or fine-tuning on object data}. As shown in Figure~\ref{fig:multiobject_general}, given a canvas with multiple object and human reference images pasted, the Artist synthesizes photorealistic outputs with natural scene composition. This is a strong result, as the model was never exposed to object-centric training data. The spatially-grounded rewards and compositional training generalize naturally to arbitrary identity-preserving generation beyond faces. Beyond simple placement, the Artist demonstrates several emergent compositional capabilities: (1) \textbf{appearance editing}, such as adding sunglasses to plush toys, with the edit naturally blended into the scene lighting and environment; (2) \textbf{object interaction}, where multiple objects relate to one another coherently within the scene; and (3) \textbf{depth ordering}, where as shown in Figure~\ref{fig:multiobject_occlusion}, our token sharing scheme with shared RoPE encodings naturally resolves spatial conflicts between overlapping objects, producing realistic occlusions and layering. These results suggest that our model learns generalizable compositional capabilities that extend well beyond the multi-human domain.

\clearpage
\section{Limitations and Future Work}
\label{app:limitations}

While Ar2Can achieves state-of-the-art performance on multi-human generation, some limitations remain for future exploration. As the number of identities scales beyond 3 people, we observe reduced control over individual facial expressions and body poses. Although our method supports prompt-based editing for fine-grained attributes such as clothing, hairstyles, and accessories across all group sizes, per-person expression control in larger groups ($>3$ people) remains challenging. Additionally, the current approach struggles with complex multi-person poses and interactions, as reflected in our Action-C scores, which show room for improvement compared to simpler action scenarios. Future work will focus on: (1) extending fine-grained expression and pose control to larger groups, (2) improving complex action generation through enhanced pose conditioning or hierarchical scene composition, and (3) exploring scalability to even larger group sizes (8+ people) while maintaining individual control. These directions will enable more versatile and controllable multi-human generation for diverse applications.

\section{Disclosure of LLM Use}
We used large language models (LLMs) for editing assistance, including grammar correction, language refinement, and vocabulary improvement. We also used LLMs for generating training prompts as described in Appendix~\ref{app:implementation}. All scientific ideas, technical contributions, experimental designs, and conclusions presented in this paper are entirely our own.

%% file: main.bib
@String(CVPR= {IEEE Conf. Comput. Vis. Pattern Recog.})

@String(ICCV= {Int. Conf. Comput. Vis.})

@String(ECCV= {Eur. Conf. Comput. Vis.})

@String(ICLR = {Int. Conf. Learn. Represent.})

@String(AAAI = {AAAI})

@String(CVPR  = {CVPR})

@String(ICCV  = {ICCV})

@String(ECCV  = {ECCV})

@String(ICLR  = {ICLR})

@inproceedings{rombach2022high,
  title={High-resolution image synthesis with latent diffusion models},
  author={Rombach, Robin and Blattmann, Andreas and Lorenz, Dominik and Esser, Patrick and Ommer, Bj{\"o}rn},
  booktitle={CVPR},
  pages={10684--10695},
  year={2022}
}

@article{ramesh2022hierarchical,
  title={Hierarchical text-conditional image generation with clip latents},
  author={Ramesh, Aditya and Dhariwal, Prafulla and Nichol, Alex and Chu, Casey and Chen, Mark},
  journal={arXiv preprint arXiv:2204.06125},
  year={2022}
}

@article{labs2025flux1kontext,
  title={FLUX. 1 Kontext: Flow Matching for In-Context Image Generation and Editing in Latent Space},
  author={Labs, Black Forest and Batifol, Stephen and Blattmann, Andreas and Boesel, Frederic and Consul, Saksham and Diagne, Cyril and Dockhorn, Tim and English, Jack and English, Zion and Esser, Patrick and others},
  journal={arXiv preprint arXiv:2506.15742},
  year={2025}
}

@article{saharia2022photorealistic,
  title={Photorealistic text-to-image diffusion models with deep language understanding},
  author={Saharia, Chitwan and Chan, William and Saxena, Saurabh and Li, Lala and Whang, Jay and Denton, Emily L and Ghasemipour, Kamyar and Gontijo Lopes, Raphael and Karagol Ayan, Burcu and Salimans, Tim and others},
  journal={NeurIPS},
  volume={35},
  pages={36479--36494},
  year={2022}
}

@inproceedings{podell2023sdxl,
  title={Sdxl: Improving latent diffusion models for high-resolution image synthesis},
  author={Podell, Dustin and English, Zion and Lacey, Kyle and Blattmann, Andreas and Dockhorn, Tim and M{\"u}ller, Jonas and Penna, Joe and Rombach, Robin},
  booktitle={ICLR},
  year={2024}
}

@article{flux2024,
  title={FLUX.1 [dev]},
  author={Black Forest Labs},
  year={2024}
}

@inproceedings{li2023gligen,
  title={Gligen: Open-set grounded text-to-image generation},
  author={Li, Yuheng and Liu, Haotian and Wu, Qingyang and Mu, Fangzhou and Yang, Jianwei and Gao, Jianfeng and Li, Chunyuan and Lee, Yong Jae},
  booktitle={CVPR},
  pages={22511--22521},
  year={2023}
}

@inproceedings{yang2023reco,
  title={Reco: Region-controlled text-to-image generation},
  author={Yang, Zhengyuan and Wang, Jianfeng and Gan, Zhe and Li, Linjie and Lin, Kevin and Wu, Chenfei and Duan, Nan and Liu, Zicheng and Liu, Ce and Zeng, Michael and others},
  booktitle={Proceedings of the IEEE/CVF Conference on Computer Vision and Pattern Recognition},
  pages={14246--14255},
  year={2023}
}

@article{ye2023ip,
  title={Ip-adapter: Text compatible image prompt adapter for text-to-image diffusion models},
  author={Ye, Hu and Zhang, Jun and Liu, Sibo and Han, Xiao and Yang, Wei},
  journal={arXiv preprint arXiv:2308.06721},
  year={2023}
}

@article{borse2025multihuman,
  title={MultiHuman-Testbench: Benchmarking Image Generation for Multiple Humans},
  author={Borse, Shubhankar and Choi, Seokeon and Park, Sunghyun and Kim, Jeongho and Kadambi, Shreya and Garrepalli, Risheek and Yun, Sungrack and Hayat, Munawar and Porikli, Fatih},
  journal={arXiv preprint arXiv:2506.20879},
  year={2025}
}

@article{qwen2024,
  title={Qwen2.5 technical report},
  author={Qwen Team},
  journal={arXiv preprint arXiv:2412.15115},
  year={2024}
}

@article{shao2024deepseekmath,
  title={Deepseekmath: Pushing the limits of mathematical reasoning in open language models},
  author={Shao, Zhihong and Wang, Peiyi and Zhu, Qihao and Xu, Runxin and Song, Junxiao and Zhang, Mingchuan and Li, YK and Wu, Yu and Guo, Daya},
  journal={arXiv preprint arXiv:2402.03300},
  year={2024}
}

@inproceedings{ma2025hpsv3,
  title={Hpsv3: Towards wide-spectrum human preference score},
  author={Ma, Yuhang and Wu, Xiaoshi and Sun, Keqiang and Li, Hongsheng},
  booktitle={Proceedings of the IEEE/CVF International Conference on Computer Vision},
  pages={15086--15095},
  year={2025}
}

@inproceedings{wu2017simultaneous,
  title={Simultaneous facial landmark detection, pose and deformation estimation under facial occlusion},
  author={Wu, Yue and Gou, Chao and Ji, Qiang},
  booktitle={Proceedings of the IEEE conference on computer vision and pattern recognition},
  pages={3471--3480},
  year={2017}
}

@inproceedings{rezatofighi2019generalized,
  title={Generalized intersection over union: A metric and a loss for bounding box regression},
  author={Rezatofighi, Hamid and Tsoi, Nathan and Gwak, JunYoung and Sadeghian, Amir and Reid, Ian and Savarese, Silvio},
  booktitle={CVPR},
  pages={658--666},
  year={2019}
}

@article{esser2024scaling,
  title={Scaling rectified flow transformers for high-resolution image synthesis},
  author={Esser, Patrick and Kulal, Sumith and Blattmann, Andreas and others},
  journal={arXiv preprint arXiv:2403.03206},
  year={2024}
}

@inproceedings{zheng2023layoutdiffusion,
  title={Layoutdiffusion: Controllable diffusion model for layout-to-image generation},
  author={Zheng, Guangcong and Zhou, Xianpan and Li, Xuewei and Qi, Zhongang and Shan, Ying and Li, Xi},
  booktitle={CVPR},
  pages={22490--22499},
  year={2023}
}

@inproceedings{xie2023boxdiff,
  title={Boxdiff: Text-to-image synthesis with training-free box-constrained diffusion},
  author={Xie, Jinheng and Li, Yuexiang and Huang, Yawen and Liu, Haozhe and Zhang, Wentian and Zheng, Yefeng and Shou, Mike Zheng},
  booktitle={ICCV},
  pages={7452--7461},
  year={2023}
}

@article{wang2024instantid,
  title={Instantid: Zero-shot identity-preserving generation in seconds},
  author={Wang, Qixun and Bai, Xu and Wang, Haofan and Qin, Zekui and Chen, Anthony},
  journal={arXiv preprint arXiv:2401.07519},
  year={2024}
}

@article{guo2024pulid,
  title={Pulid: Pure and lightning id customization via contrastive alignment},
  author={Guo, Zinan and Wu, Yanze and Chen, Zhuowei and Chen, Lang and He, Qian},
  journal={arXiv preprint arXiv:2404.16022},
  year={2024}
}

@article{yang2024consistentid,
  title={Consistentid: Portrait generation with multimodal fine-grained identity preserving},
  author={Huang, Jiehui and Dong, Xiao and Song, Wenhui and Chong, Zheng and Tang, Zhenchao and Zhou, Jun and Cheng, Yuhao and Chen, Long and Li, Hanhui and Yan, Yiqiang and others},
  journal={IEEE Transactions on Pattern Analysis and Machine Intelligence},
  year={2026},
  publisher={IEEE}
}

@article{black2023training,
  title={Training diffusion models with reinforcement learning},
  author={Black, Kevin and Janner, Michael and Du, Yilun and Kostrikov, Ilya and Levine, Sergey},
  journal={arXiv preprint arXiv:2305.13301},
  year={2023}
}

@article{xu2023imagereward,
  title={Imagereward: Learning and evaluating human preferences for text-to-image generation},
  author={Xu, Jiazheng and Liu, Xiao and Wu, Yuchen and Tong, Yuxuan and Li, Qinkai and Ding, Ming and Tang, Jie and Dong, Yuxiao},
  journal={NeurIPS},
  volume={36},
  year={2023}
}

@article{fan2023dpok,
  title={Dpok: Reinforcement learning for fine-tuning text-to-image diffusion models},
  author={Fan, Ying and Watkins, Olivia and Du, Yuqing and Liu, Hao and Ryu, Moonkyung and Boutilier, Craig and Abbeel, Pieter and Ghavamzadeh, Mohammad and Lee, Kangwook and Lee, Kimin},
  journal={Advances in Neural Information Processing Systems},
  volume={36},
  pages={79858--79885},
  year={2023}
}

@article{prabhudesai2023aligning,
  title={Aligning text-to-image diffusion models with reward backpropagation},
  author={Prabhudesai, Mihir and Goyal, Anirudh and Pathak, Deepak and Fragkiadaki, Katerina},
  journal={arXiv preprint arXiv:2310.03739},
  year={2023}
}

@article{disco2025,
  title={DisCo: Reinforcement with Diversity Constraints for Multi-Human Generation},
  author={Borse, Shubhankar and Farhadzadeh, Farzad and Hayat, Munawar and Porikli, Fatih},
  journal={arXiv preprint arXiv:2510.01399},
  year={2025}
}

@inproceedings{deng2019arcface,
  title={Arcface: Additive angular margin loss for deep face recognition},
  author={Deng, Jiankang and Guo, Jia and Xue, Niannan and Zafeiriou, Stefanos},
  booktitle={CVPR},
  pages={4690--4699},
  year={2019}
}

@inproceedings{deng2020retinaface,
  title={Retinaface: Single-shot multi-level face localisation in the wild},
  author={Deng, Jiankang and Guo, Jia and Ververas, Evangelos and Kotsia, Irene and Zafeiriou, Stefanos},
  booktitle={CVPR},
  pages={5203--5212},
  year={2020}
}

@article{kuhn1955hungarian,
  title={The Hungarian method for the assignment problem},
  author={Kuhn, Harold W},
  journal={Naval Research Logistics Quarterly},
  volume={2},
  pages={83--97},
  year={1955}
}

@inproceedings{chefer2023attend,
  title={Attend-and-excite: Attention-based semantic guidance for text-to-image diffusion models},
  author={Chefer, Hila and Alaluf, Yuval and Vinker, Yael and Wolf, Lior and Cohen-Or, Daniel},
  booktitle={SIGGRAPH},
  pages={1--10},
  year={2023}
}

@inproceedings{liu2022compositional,
  title={Compositional visual generation with composable diffusion models},
  author={Liu, Nan and Li, Shuang and Du, Yilun and Torralba, Antonio and Tenenbaum, Joshua B},
  booktitle={ECCV},
  pages={423--439},
  year={2022}
}

@inproceedings{lisa,
  title={Lisa: Reasoning segmentation via large language model},
  author={Lai, Xin and Tian, Zhuotao and Chen, Yukang and Li, Yanwei and Yuan, Yuhui and Liu, Shu and Jia, Jiaya},
  booktitle={Proceedings of the IEEE/CVF Conference on Computer Vision and Pattern Recognition},
  pages={9579--9589},
  year={2024}
}

@inproceedings{chatpose,
  title={Chatpose: Chatting about 3d human pose},
  author={Feng, Yao and Lin, Jing and Dwivedi, Sai Kumar and Sun, Yu and Patel, Priyanka and Black, Michael J},
  booktitle={Proceedings of the IEEE/CVF conference on computer vision and pattern recognition},
  pages={2093--2103},
  year={2024}
}

@inproceedings{chatgarment,
  title={Chatgarment: Garment estimation, generation and editing via large language models},
  author={Bian, Siyuan and Xu, Chenghao and Xiu, Yuliang and Grigorev, Artur and Liu, Zhen and Lu, Cewu and Black, Michael J and Feng, Yao},
  booktitle={Proceedings of the Computer Vision and Pattern Recognition Conference},
  pages={2924--2934},
  year={2025}
}

@inproceedings{he2025uniportrait,
  title={Uniportrait: A unified framework for identity-preserving single-and multi-human image personalization},
  author={He, Junjie and Geng, Yifeng and Bo, Liefeng},
  booktitle={Proceedings of the IEEE/CVF International Conference on Computer Vision},
  pages={14399--14408},
  year={2025}
}

@inproceedings{zhang2025id,
  title={Id-patch: Robust id association for group photo personalization},
  author={Zhang, Yimeng and Zhi, Tiancheng and Liu, Jing and Sang, Shen and Jiang, Liming and Yan, Qing and Liu, Sijia and Luo, Linjie},
  booktitle={Proceedings of the Computer Vision and Pattern Recognition Conference},
  pages={2986--2996},
  year={2025}
}

@inproceedings{xiao2025omnigen,
  title={Omnigen: Unified image generation},
  author={Xiao, Shitao and Wang, Yueze and Zhou, Junjie and Yuan, Huaying and Xing, Xingrun and Yan, Ruiran and Li, Chaofan and Wang, Shuting and Huang, Tiejun and Liu, Zheng},
  booktitle={Proceedings of the Computer Vision and Pattern Recognition Conference},
  pages={13294--13304},
  year={2025}
}

@inproceedings{aipparel,
  title={AIpparel: A Multimodal Foundation Model for Digital Garments},
  author={Nakayama, Kiyohiro and Ackermann, Jan and Kesdogan, Timur Levent and Zheng, Yang and Korosteleva, Maria and Sorkine-Hornung, Olga and Guibas, Leonidas J and Yang, Guandao and Wetzstein, Gordon},
  booktitle={Proceedings of the Computer Vision and Pattern Recognition Conference},
  pages={8138--8149},
  year={2025}
}

@article{wu2025omnigen2,
  title={OmniGen2: Exploration to Advanced Multimodal Generation},
  author={Wu, Chenyuan and Zheng, Pengfei and Yan, Ruiran and Xiao, Shitao and Luo, Xin and Wang, Yueze and Li, Wanli and Jiang, Xiyan and Liu, Yexin and Zhou, Junjie and others},
  journal={arXiv preprint arXiv:2506.18871},
  year={2025}
}

@article{mou2025dreamo,
  title={Dreamo: A unified framework for image customization},
  author={Mou, Chong and Wu, Yanze and Wu, Wenxu and Guo, Zinan and Zhang, Pengze and Cheng, Yufeng and Luo, Yiming and Ding, Fei and Zhang, Shiwen and Li, Xinghui and others},
  journal={arXiv preprint arXiv:2504.16915},
  year={2025}
}

@article{chen2025xverse,
  title={XVerse: Consistent Multi-Subject Control of Identity and Semantic Attributes via DiT Modulation},
  author={Chen, Bowen and Zhao, Mengyi and Sun, Haomiao and Chen, Li and Wang, Xu and Du, Kang and Wu, Xinglong},
  journal={arXiv preprint arXiv:2506.21416},
  year={2025}
}

@article{xu2025withanyone,
  title={WithAnyone: Towards Controllable and ID Consistent Image Generation},
  author={Xu, Hengyuan and Cheng, Wei and Xing, Peng and Fang, Yixiao and Wu, Shuhan and Wang, Rui and Zeng, Xianfang and Jiang, Daxin and Yu, Gang and Ma, Xingjun and others},
  journal={arXiv preprint arXiv:2510.14975},
  year={2025}
}

@article{cheng2025umo,
  title={UMO: Scaling Multi-Identity Consistency for Image Customization via Matching Reward},
  author={Cheng, Yufeng and Wu, Wenxu and Wu, Shaojin and Huang, Mengqi and Ding, Fei and He, Qian},
  journal={arXiv preprint arXiv:2509.06818},
  year={2025}
}

@inproceedings{guo2016ms,
  title={Ms-celeb-1m: A dataset and benchmark for large-scale face recognition},
  author={Guo, Yandong and Zhang, Lei and Hu, Yuxiao and He, Xiaodong and Gao, Jianfeng},
  booktitle={European conference on computer vision},
  pages={87--102},
  year={2016},
  organization={Springer}
}

@article{liu2025flow,
  title={Flow-grpo: Training flow matching models via online rl},
  author={Liu, Jie and Liu, Gongye and Liang, Jiajun and Li, Yangguang and Liu, Jiaheng and Wang, Xintao and Wan, Pengfei and Zhang, Di and Ouyang, Wanli},
  journal={arXiv preprint arXiv:2505.05470},
  year={2025}
}

@article{deng2025emerging,
  title={Emerging properties in unified multimodal pretraining},
  author={Deng, Chaorui and Zhu, Deyao and Li, Kunchang and Gou, Chenhui and Li, Feng and Wang, Zeyu and Zhong, Shu and Yu, Weihao and Nie, Xiaonan and Song, Ziang and others},
  journal={arXiv preprint arXiv:2505.14683},
  year={2025}
}

@article{qian2025layercomposer,
  title={LayerComposer: Multi-Human Personalized Generation via Layered Canvas},
  author={Qian, Guocheng Gordon and Zhang, Ruihang and Chen, Tsai-Shien and Dalva, Yusuf and Goyal, Anujraaj Argo and Menapace, Willi and Skorokhodov, Ivan and Dong, Meng and Sahni, Arpit and Ostashev, Daniil and others},
  journal={arXiv preprint arXiv:2510.20820},
  year={2025}
}

@article{dalva2025canvas,
  title={Canvas-to-Image: Compositional Image Generation with Multimodal Controls},
  author={Dalva, Yusuf and Qian, Guocheng Gordon and Goldenberg, Maya and Chen, Tsai-Shien and Aberman, Kfir and Tulyakov, Sergey and Yanardag, Pinar and Wang, Kuan-Chieh Jackson},
  journal={arXiv preprint arXiv:2511.21691},
  year={2025}
}

@misc{qian2025composeme,
      title={ComposeMe: Attribute-Specific Image Prompts for Controllable Human Image Generation}, 
      author={Guocheng Gordon Qian and Daniil Ostashev and Egor Nemchinov and Avihay Assouline and Sergey Tulyakov and Kuan-Chieh Jackson Wang and Kfir Aberman},
      year={2025},
      eprint={2509.18092},
      archivePrefix={arXiv},
      primaryClass={cs.CV},
      url={https://arxiv.org/abs/2509.18092}, 
}

@inproceedings{qian2025omni,
  title={Omni-id: Holistic identity representation designed for generative tasks},
  author={Qian, Guocheng and Wang, Kuan-Chieh and Patashnik, Or and Heravi, Negin and Ostashev, Daniil and Tulyakov, Sergey and Cohen-Or, Daniel and Aberman, Kfir},
  booktitle={Proceedings of the IEEE/CVF Conference on Computer Vision and Pattern Recognition},
  pages={8786--8795},
  year={2025}
}

@inproceedings{huang2025resolving,
  title={Resolving multi-condition confusion for finetuning-free personalized image generation},
  author={Huang, Qihan and Fu, Siming and Liu, Jinlong and Jiang, Hao and Yu, Yipeng and Song, Jie},
  booktitle={Proceedings of the AAAI Conference on Artificial Intelligence},
  volume={39},
  number={4},
  pages={3707--3714},
  year={2025}
}

@article{feng2023layoutgpt,
  title={Layoutgpt: Compositional visual planning and generation with large language models},
  author={Feng, Weixi and Zhu, Wanrong and Fu, Tsu-jui and Jampani, Varun and Akula, Arjun and He, Xuehai and Basu, Sugato and Wang, Xin Eric and Wang, William Yang},
  journal={Advances in Neural Information Processing Systems},
  volume={36},
  pages={18225--18250},
  year={2023}
}

@inproceedings{yang2024mastering,
  title={Mastering Text-to-Image Diffusion: Recaptioning, Planning, and Generating with Multimodal LLMs},
  author={Yang, Ling and Yu, Zhaochen and Meng, Chenlin and Xu, Minkai and Ermon, Stefano and Cui, Bin},
  booktitle={International Conference on Machine Learning},
  year={2024}
}

@article{zhou2024storymaker,
  title={Storymaker: Towards holistic consistent characters in text-to-image generation},
  author={Zhou, Zhengguang and Li, Jing and Li, Huaxia and Chen, Nemo and Tang, Xu},
  journal={arXiv preprint arXiv:2409.12576},
  year={2024}
}

@inproceedings{jiang2025infiniteyou,
  title={Infiniteyou: Flexible photo recrafting while preserving your identity},
  author={Jiang, Liming and Yan, Qing and Jia, Yumin and Liu, Zichuan and Kang, Hao and Lu, Xin},
  booktitle={Proceedings of the IEEE/CVF International Conference on Computer Vision},
  pages={10898--10907},
  year={2025}
}
